\documentclass{article}

\PassOptionsToPackage{numbers, compress}{natbib}

\usepackage{natbib}

\usepackage[final]{neurips_2022}

\usepackage[utf8]{inputenc} 
\usepackage[T1]{fontenc}    
\usepackage{url}            
\usepackage{booktabs}       
\usepackage{amsfonts}       
\usepackage{nicefrac}       
\usepackage{microtype}      
\usepackage{xcolor}         
\usepackage{enumitem}
\usepackage{multirow}
\usepackage{tabularx}
\usepackage{amsmath}
\usepackage{nicefrac}
\usepackage{adjustbox}
\usepackage{caption}
\usepackage{subcaption}
\usepackage{soul}
\usepackage{wrapfig}

\definecolor{applegreen}{rgb}{0.55, 0.71, 0.0}
\definecolor{aogreen}{rgb}{0.0, 0.5, 0.0}

\newcommand\yeon[1]{\textcolor{black}{#1}}

\newcommand{\halluNE}{$\textsc{NE}_{\textsc{Er}}$}
\newcommand{\entailedRatio}{$\text{Entail}_{\textsc{R}}$}
\newcommand{\distFour}{Div.}
\newcommand{\prompts}{\textsc{FactualityPrompts}}
\newcommand{\prefix}{\textsc{TopicPrefix}}

\newcommand{\halfSC}{$SC_{\textsc{half}}$}
\newcommand{\rootSC}{$SC_{\textsc{root}}$}
\newcommand{\randomSC}{$SC_{\textsc{random}}$}
\newcommand{\neSC}{$SC_{\textsc{NE}}$}
\newcommand{\reset}{$p$-reset}
\newcommand{\lowerBound}{$\omega$-bound}
\newcommand{\decay}{$\lambda$-decay}

\definecolor{darker}{rgb}{0,0.15,0.7}
\usepackage[colorlinks, urlcolor=darker, citecolor=darker, linkcolor=darker]{hyperref}

\title{Factuality Enhanced Language Models  for \\  Open-Ended Text Generation}
\usepackage{authblk}

\author[1]{Nayeon Lee\thanks{\noindent  Work done during an internship at NVIDIA.}\ \ $^\dagger$}
\affil[1]{Hong Kong University of Science and Technology}
\affil[2]{NVIDIA} 

\author[2]{Wei Ping\thanks{Correspondence to: Nayeon Lee <nayeon.lee@connect.ust.hk>, Wei Ping <wping@nvidia.com>.}\ \ }
\author[2]{Peng Xu}
\author[2]{Mostofa Patwary}
\author[1]{Pascale Fung}
\author[2]{\\ \vspace{1mm}Mohammad Shoeybi}
\author[2]{Bryan Catanzaro}

\begin{document}

\maketitle

\begin{abstract}
Pretrained language models~(LMs) are susceptible to generate text with nonfactual information.
In this work, we measure and improve the factual accuracy of large-scale LMs for open-ended text generation.
We design the \prompts~test set and metrics to measure the factuality of LM generations. 
Based on that, we study the factual accuracy of LMs with parameter sizes ranging from 126M to 530B. 
Interestingly, we find that larger LMs are more \textit{factual} than smaller ones, although a previous study suggests that larger LMs can be less truthful in terms of \textit{misconceptions}.
In addition, popular sampling algorithms~(e.g., top-$p$) in open-ended text generation can harm the factuality due to the ``uniform randomness'' introduced at every sampling step.
We propose the \emph{factual-nucleus} sampling algorithm that dynamically adapts the randomness to improve the factuality of generation while maintaining quality.
Furthermore, we analyze the inefficiencies of the standard training method in learning correct associations between entities from factual text corpus~(e.g., Wikipedia). 
We propose a \emph{factuality-enhanced} training method that uses 
{\prefix} for better awareness of facts and sentence completion as the training objective, which can vastly reduce the factual errors.
We release our code and \prompts~benchmark at:  \url{https://github.com/nayeon7lee/FactualityPrompt}.
\end{abstract}

\vspace{-4mm}
\section{Introduction}
\label{sec:introduction}
\vspace{-2mm}
Large-scale pre-trained language models (LMs) have demonstrated impressive natural language generation results~\citep{radford2019language, raffel2019exploring, brown2020language, smith2022using}. 
However, the generative LMs~(e.g., GPT-3) 
are solely trained to model the statistical correlations between subword tokens~\citep{sennrich2015neural}, and have limited capability to 
generate factually accurate text 
as illustrated in Table~\ref{table:illustration}.
As a result, there are increasing concerns about the nonfactual generations from large-scale pre-trained LMs~\citep[e.g.,][]{rae2021scaling, nakano2021webgpt,zhang2022opt}, which needs to be adequately addressed for their safe deployment in real-world applications, e.g., content creation~\citep{zellers2019defending} and dialogue~\citep{thoppilan2022lamda}.

In previous studies, different metrics and methods have been proposed to measure and improve the factual accuracy of language generation within different tasks~\citep{ji2022survey}, including text summarization~\citep[e.g.,][]{kryscinski2019evaluating, maynez2020faithfulness, durmus2020feqa, nan2021improving}, question answering~\cite[e.g.,][]{yin2015neural, roberts2020much, su2022read}, and table-to-text generation~\citep[e.g.,][]{moryossef2019step, liu2021towards}.
However, these works focus on the faithfulness (or factuality) of the \textit{fine-tuned} LMs for particular downstream tasks~(i.e., factual consistency between source and target text). Little exploration has been made to address the factual errors in \textit{pretrained} LMs for general-purpose open-ended text generation, where the goal is to generate a coherent continuation from the given context~(e.g., the use cases from GPT-2). 

One of the popular methods for enhancing generation factuality is to incorporate external knowledge sources~\citep{yu2020survey,piktus2021web,west2022probing}.
Structured knowledge bases and graphs have been utilized for grounded text generation~\citep[e.g.,][]{ahn2016neural, logan2019barack}, where the LMs are trained to select and copy relevant facts from external knowledge sources.
In contrast to the sizeable online text with factual information, the structured knowledge graphs only encode a limited amount of knowledge as they require expensive human annotations for high-quality construction.
A method that can directly leverage plain text knowledge~(e.g., Wikipedia,  encyclopedia books, peer-reviewed publications) would be desirable for factuality enhancement as it can remove the human annotation bottleneck and easily scale up the amount of injected knowledge.
Augmenting LM with an information retrieval (IR) system is one possible solution to leverage textual facts, however, at the cost of additional complexity and resource overhead to the model~\citep{thoppilan2022lamda, borgeaud2021improving, piktus2021web, petroni2020kilt, guu2020realm}.
Therefore, we explore an IR-free method that enhances the innate factuality of LMs by
continued training on a factually rich plain-text corpus.

In this work, we focus on measuring and improving the factuality of large-scale pre-trained language models~(LMs) for open-ended text generation.
Specifically, we make the following contributions: 
\vspace{-2mm}
\begin{enumerate}[leftmargin=3.1em]
    \item We build the benchmark and metrics to measure the factual accuracy of pre-trained LM for open-ended text generation. \
    We demonstrate a good correlation between the proposed automatic metrics and human assessment of factuality.
    Based on that, we systematically study the factual accuracy of LMs with parameter sizes ranging from 126M to 530B and find that large LMs have higher factual accuracy than smaller ones~(e.g., named-entity factual error is reduced from 63.69\% to 33.3\%).
    \vspace{-1mm}
    \item We study the decoding algorithms of LM in terms of factual accuracy. We unveil that the popular nucleus sampling algorithm~\citep{holtzman2019curious} for open-ended text generation can easily mix up different named entities or randomly fabricate information due to the ``uniform randomness'' introduced at every decoding step. 
    We propose \emph{factual-nucleus} sampling algorithm that promotes generation factuality while maintaining the quality and diversity.
    \vspace{-1mm}
    \item We explore training methods that can effectively leverage text corpus with rich facts~(e.g., Wikipedia). We find that directly continuing the training of LM on factual text data~\cite{gururangan2020don} does not guarantee the improvement of factual accuracy. 
    We propose \emph{factuality-enhanced} training to address the underlying inefficiencies of this baseline. Our method consists of i) an addition of a {\prefix} that improves the awareness of facts during training, and ii) a sentence completion task as the new objective for continued LM training~\citep[e.g.,][]{gururangan2020don}.   
    \vspace{-1mm}
    \item We demonstrate that the factual accuracy of large-scale LMs~(up to 530B) can be significantly enhanced~(i.e., named-entity factual error is reduced from 33.3\% to 14.5\%) after applying the proposed \emph{factuality-enhanced} training with \emph{factual-nucleus} sampling algorithm. 
  \vspace{-1mm}
\end{enumerate}

We organize the rest of the paper as follows.
We discuss related work in \S~\ref{sec:related_work} and present our benchmark setup with evaluation protocol in \S~\ref{sec:benchmark}.
We study the factual accuracy of LMs with respect to model size, prompt type, and choice of decoding algorithm in \S~\ref{sec:case_study}.
After that, we present \emph{factual-nucleus} sampling algorithm in \S~\ref{sec:dynamic_topp}, and \emph{factuality-enhanced} training in \S~\ref{sec:factual_training}.
We conclude the paper in \S~\ref{sec:conclusion}.

\begin{table}[]
\vspace{-4mm}
\centering
\caption{\footnotesize
Example of continuations from the 530B LM with greedy decoding and nucleus sampling $p=0.9$, along with the continuation from factuality-enhanced 530B LM with \emph{factual-nucleus sampling}.
\textcolor{red}{Red} represents nonfactual, \textcolor{aogreen}{green} represents factual, and \st{strikethrough} represents repetition. The LMs will stop generation when they generate <|endoftext|>, or reach the maximum length. See Appendix~\ref{appendix:more_samples} for more examples.
}
\vspace{1mm}
\small
\label{table:illustration}
\begin{tabular}{p{0.95\linewidth}}
\toprule
\textbf{[Factual Prompt]} One of lung cancer's symptoms is shortness of breath. \\ \midrule
\textbf{[Greedy]} \textcolor{aogreen}{The American Cancer Society estimates that more than 228,000 people will be diagnosed with lung cancer in the United States this year.\textbackslash{}n\textbackslash{}n}
\textcolor{aogreen}{
\st{The American Cancer Society estimates that more than 228,000 people will be diagnosed with lung cancer in the United States this year.\textbackslash{}n\textbackslash{}nThe American Cancer Society estimates that more than 228,000 people will be diagnosed with lung cancer in the United...}} \\
\vspace{0mm}
\textbf{[p=0.9]} \textcolor{red}{That would make an oxygen mask one of the more popular treatments for this devastating disease.} \textcolor{aogreen}{It helps ease breathing and give patients back their strength.} Learn more here.<|endoftext|>" \\
\vspace{0mm}
\vspace{0mm}
\textbf{[Ours]} \textcolor{aogreen}{Other symptoms include coughing (including coughing up blood), chest pain, \textcolor{red}{hoarseness}, weight loss, and loss of appetite. The symptoms of lung cancer may resemble those of a bad cold or pneumonia. A chest x - ray is \textcolor{red}{necessary} to confirm the diagnosis. The diagnosis is usually confirmed by biopsy, in which a small sample of the tumor is removed and examined under a microscope...} 
\\ 
\midrule
\textbf{[Non-Factual Prompt]} Cancer can be prevented by drinking too much alcohol. \\ \midrule
\textbf{[Greedy]} \textcolor{red}{\textbackslash{}nCancer can be prevented by eating too much red meat.\textbackslash{}n\textbackslash{}n\st{Cancer can be prevented by eating too much} sugar.\st{\textbackslash{}n\textbackslash{}nCancer can be prevented by eating too much} fat.\st{\textbackslash{}n\textbackslash{}nCancer can be prevented by eating too much} processed food.\st{\textbackslash{}n\textbackslash{}nCancer can be}...} \\
\vspace{0mm}
\textbf{[p=0.9]} \textcolor{red}{More than 90\% of hepatocellular carcinoma is associated with alcohol use.\textbackslash{}n\textbackslash{}nAbout 9 percent of cancers can be prevented if we don't eat more than 1.5 ounces of processed meat each day.\textbackslash{}n\textbackslash{}nDiet is believed to play a role in 7\% of cancers and...} \\
\vspace{0mm}
\textbf{[Ours]} \textcolor{aogreen}{Cancer is often treated with some combination of radiation therapy, surgery, chemotherapy and targeted therapy. Pain and symptom management are an important part of care. Palliative care is particularly important in people with advanced disease. The chance of survival depends on the type of cancer and extent of disease at the...}  \\
\bottomrule
\end{tabular}
\vspace{-1mm}
\end{table}

\vspace{-1mm}
\section{Related Work}
\label{sec:related_work}
\vspace{-1mm}
\textbf{Factuality vs. Model Size}~\ 
\citet{lin2021truthfulqa} propose the TruthfulQA benchmark to measure the falsehood generations from different sized LMs. The result suggests that bigger LMs pre-trained on web text are generally less truthful than smaller ones in terms of false belief or misconception. At first glance, this is contradictory to our observation, however, our work focuses on different knowledge to TruthfulQA work. The TruthfulQA benchmark focuses on conceptual knowledge, while our benchmark focuses on factual knowledge~\citep{krathwohl2002revision}~\footnote{According to \citet{krathwohl2002revision},  knowledge can be categorized into four types: i) factual knowledge, ii) conceptual knowledge, iii) procedural knowledge, and iv) metacognitive knowledge.}.
Large LMs can be good at recalling factual knowledge given substantial pre-training corpus, suggested by previous studies on LM's parameteric knowledge~\citep{petroni2019language}, but there still remains room for improvement for reasoning conceptual knowledge~\citep{aspillaga2021inspecting, zhou2020evaluating}.

\textbf{Parametric Factual Knowledge}~\
A group of work addresses the factual errors in the parametric knowledge of LMs that is acquired from training corpus~\citep{jiang2020can, zhong2021factual,elazar2021measuring}.
The correctness of the parametric knowledge is commonly tested in cloze-style question answering format~\citep{petroni2019language} (e.g., Person X is born in \_\_). Efforts are made to fine-tune the pre-trained LM to ``inject'' more knowledge and improve its ability to answer factual questions without consulting external knowledge source~\citep{roberts2020much}.  Moreover, some works attempt to edit and fix the factual errors~\citep{de2021editing,jang2021towards,meng2022locating}. However, it is unclear if the improvement of fine-tuned LM for QA-style task can help to mitigate factual errors in open-ended text generation task. 

\textbf{Hallucination in downstream NLG tasks}~\
There are active efforts to reduce the unfaithfulness or factual errors of task-specific LMs fine-tuned for various downstream natural language generation~(NLG) tasks such as summarization~\citep{cao2018faithful,dong2020multi,huang2020knowledge,huang2021factual,cao2021cliff,zhu2021enhancing,chen2021improving}, data-to-text~\citep{wiseman2017challenges,nie2019simple,liu2021towards,su2021plan,wang2021sketch,rebuffel2022controlling} and dialogue system~\citep{shen2021identifying,shuster2021retrieval,rashkin2021increasing,wu2021controllable,dziri2021neural}. In contrast to these works, we focus on general purpose LM for open-ended text generation task.

\textbf{Human-in-the-loop}~\
Human feedback or demonstrations are valuable to improve the factual accuracy of LMs. For example, InstructGPT~\citep{ouyang2022training} fine-tune the LMs with collected human feedback for a truthful generation.
WebGPT~\citep{nakano2021webgpt} is trained to cite its sources when it generates output, thus allowing humans to evaluate factual accuracy by checking whether a claim is supported by a reliable source.
In this work, we focus on human-free solution to mitigate nonfactual generations, as it is less expensive and easy to scale.

\vspace{-1mm}
\section{\prompts~and Evaluation Metrics}
\label{sec:benchmark}
\vspace{-1mm}
%
Our goal is to automatically measure and evaluate the factuality of large-scale pre-trained language models~(LMs) for open-ended text generation. Factuality refers to being coherent to provided ground-truth knowledge sources in NLP~\citep{ji2022survey}. 
The biggest challenge of evaluating factuality for open-ended text generation is associated with locating the ground-truth knowledge from the myriad of world knowledge.
Evaluating open-ended text generation can be challenging due to the lack of ground-truth references for generation~\citep{holtzman2019curious, pillutla2021mauve}.
In this study, the scope of our ground-truth knowledge source is set to Wikipedia~\footnote{Note that Wikipedia is one of the most commonly-used, accessible, large-scale, good quality, unstructured knowledge sources. 
Our proposed methods can easily generalize to other knowledge sources in plain text~(e.g., arXiv papers, medical reports, reliable newspapers).}
because this helps simplify the evaluation setup.

\begin{figure}[t]
\centering
\includegraphics[width=0.95\linewidth]{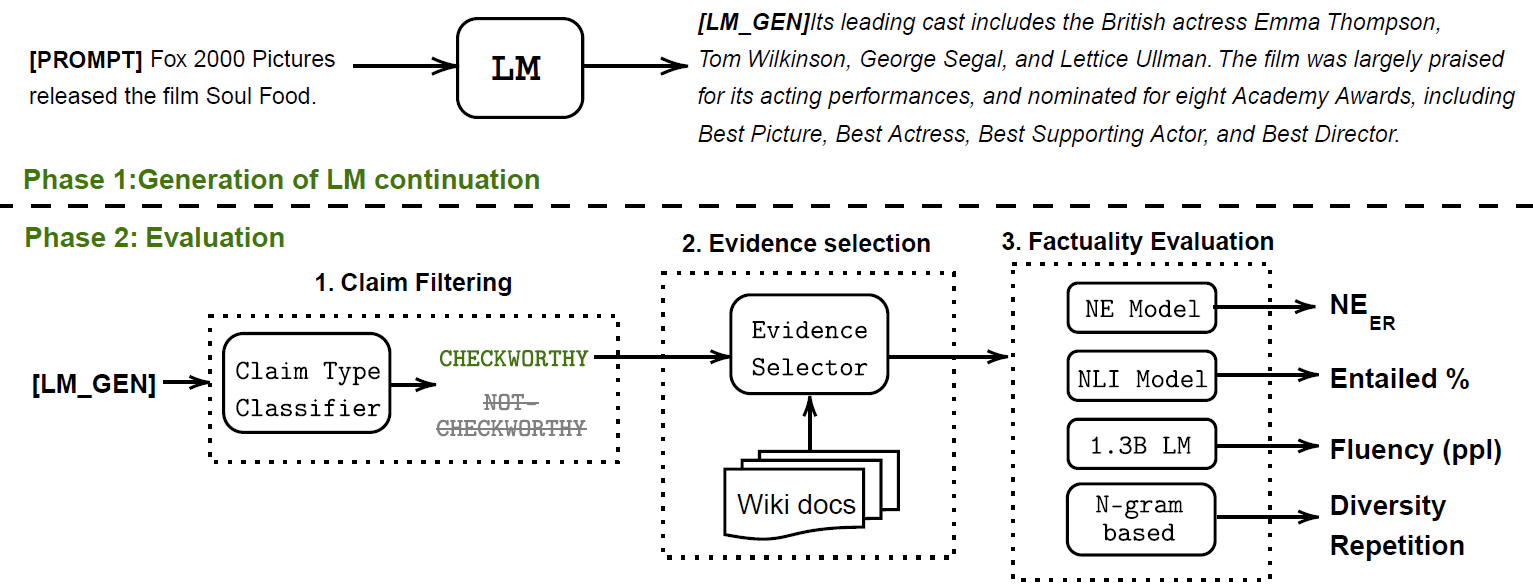}
\vspace{-2mm}
\caption{\footnotesize
Illustration of our evaluation framework}
\label{fig:evaluation_framework_illustration}
\vspace{-2mm}
\end{figure}

As illustrated in Fig~\ref{fig:evaluation_framework_illustration}, our evaluation framework consists of the following phases.
In phase 1, LM generates the continuations from the provided test prompts (\S\ref{subsec:prompts}). In phase 2, we first identify \emph{check-worthy} continuations, which refers to the generations with facts that require factuality evaluation.
One may refer to Appendix~\ref{appendix:claim_filtering} for details.
This step is necessary as open-ended text generation may generate text that does not contain facts such as personal opinion or chitchat-style text (e.g., ``I like eating apples!'').
Then, we prepare relevant ground-truth knowledge required for factual verification of \emph{check-worthy} continuations (\S\ref{subsec:knowledge_prep}). 
Lastly, we calculate the factuality and quality measures (\S\ref{subsec:metric}).

\vspace{-1mm}
\subsection{\prompts~Testset}
\label{subsec:prompts}
\vspace{-1mm}
We design our test prompts (\prompts) that follows a similar setup as in RealToxicityPrompts~\citep{gehman2020realtoxicityprompts}, which has \textit{toxic} and \textit{nontoxic} prompts to evaluate the toxicity of LM continuations. 
\prompts~consists of \textit{factual} and \textit{nonfactual} prompts that allow us to study the impact of prompts' factuality on the LM continuation; this simulates the real-world scenario where input texts are not guaranteed to be factual. \yeon{The data construction and statistic details are provided in Appendix~\ref{appendix:data_details}}, and we will release the constructed \prompts~for future research.

\vspace{-1mm}
\subsection{Ground-Truth Knowledge Preparation}
\label{subsec:knowledge_prep}
\vspace{-1mm}
To evaluate the factuality of a given generation, we need to prepare relevant ground-truth knowledge. 
The required ground-truth knowledge can be either document-level or sentence-level, depending on the type of factuality metrics (discussed in \S\ref{subsec:metric}).
The correctness of factuality evaluation is crucially dependent on the correctness of the ground-truth knowledge. To ensure that our factuality evaluation is not distorted by the irrelevant provision of ground-truth knowledge, we do the following:

For \textbf{document-level} ground-truth knowledge, we directly use the Wikipedia document annotation from the FEVER dataset. This way, we can mitigate any potential error from automatic document retrieval. 
For \textbf{sentence-level} ground-truth knowledge, we do automatic sentence selection by using two different methods to maximize the chance of recalling the relevant ground-truth knowledge.
We treat the generated text as query $q$ and Wikipedia sentences as a pool of candidates $C=\{c_1, c_2, c_3, ... c_N\}$ where $N$ is the number of sentences in the Wikipedia document.
One ground-truth sentence is retrieved by obtaining TF-IDF vector representations of $q$ and $C$ and selecting the $c_i$ with the highest cosine similarity with the $q$. Another is retrieved by obtaining the contextual representation of $q$ and $C$ using SentenceTransformer~\citep{reimers2019sentence} and selecting the $c_j$ with the highest cosine similarity. 

\vspace{-1mm}
\subsection{Evaluation Metrics}
\label{subsec:metric}
\vspace{-1mm}
We adapt commonly used metric designs from the hallucination literature~\cite{ji2022survey}:
named-entity~(NE) based metric and textual entailment based metric. Each metric captures a different aspect of factuality, so we use both metrics for better understanding of factuality. 

\vspace{-3mm}
\paragraph{Hallucinated NE Error} Since NEs are one of the core building blocks of ``fact'', NE-related metric design is one of the common choices in literature~\citep{ji2022survey,goodrich2019assessing,nan2021entity}. In this work, we specifically adopt the NE-based metric~\citep{nan2021entity} that is designed with a belief that a model is hallucinating (making factual errors) if it generates a NE that does not appear in the ground-truth knowledge source.

We define our NE-based metric to be:
$
\text{\halluNE} = {|\textsc{Hallu}_{\text{NE}}|} ~/~ {|\textsc{All}_{\text{NE}}|}
$
where $\textsc{All}_{\text{NE}}$ is the set of all the NEs detected in the LM generation, and $\textsc{Hallu}_{\text{NE}}$ is subset of $\textsc{NE}_{\text{All}}$ that does not appear in the ground-truth Wikipedia document. Note that evaluating \halluNE~requires document-level ground-truth.
%
To ensure the quality of the metric, we also take the same precautions used by \cite{nan2021entity}. For named entities consisting of multiple words, partial n-gram overlaps are also treated as a ``match''. This ensures we can address the shortened form of named entities -- e.g., ``Barack Hussein Obama II'' vs. ``Obama''. Note that stopwords (e.g., the, a) are not considered in the partial n-gram overlaps. The named entities are detected using a publicly available pre-trained NE detection model from \emph{Spacy.io}.

\vspace{-3mm}
\paragraph{Entailment Ratio}
Textual Entailment (or natural language inference) is a task of determining whether a hypothesis is \emph{entailed} by, \emph{refuted} by, or \emph{neutral} to a given premise~\citep{maccartney-manning-2008-modeling}.
Entailment-based metrics are based on the rationale that factual generation will be entailed by the ground-truth knowledge~\citep{ji2022survey,kryscinski2019evaluating,falke2019ranking,duvsek2020evaluating,dziri2021evaluating}.

We define the entailment ratio as: 
\entailedRatio $= | \textsc{Entail}_{\text{gen}}| ~/~ {|\textsc{All}_{\text{gen}}|},$
where $\textsc{All}_{\text{gen}}$ is set of all generations, and $\textsc{Entail}_{\text{gen}}$ is the set of generations that are entailed by a entailment model. To obtain the entailment scores, we leverage a pretrained entailment model that is publicly available~\footnote{Refer to the code snippet provided in \url{https://pytorch.org/hub/pytorch_fairseq_roberta/}}; a RoBERTa~\citep{liu2019roberta} model fine-tuned on MNLI~\citep{williams2017broad} dataset.
\entailedRatio~requires sentence-level ground-truth because 
only a few Wikipedia sentences are relevant to specific factual information in a given generation. For example, ``Barack Obama was born in Hawaii'' is only relevant to the Wikipedia sentence that mentions his birth location. 
{Note that our \entailedRatio~is a stricter form of metric that does not treat \emph{neutral} class to be factual.}

\vspace{-3mm}
\paragraph{Generation Quality Evaluation} 
We also evaluate the generation quality from three aspects:
\emph{i)}~\textit{Fluency} is an important aspect of text generation. We measured it by the mean perplexity of generated continuations evaluated 
{with a large pretrained LM, which is 1.3B LM in this work .}
\emph{ii)}~\textit{Diversity} is an important characteristic of LM that makes the generation more interesting and engaging -- it is bland and boring to always generate same texts. 
It is measured using the mean number of distinct n-grams (we report 4-gram), normalized by the length of text~\citep{li2015diversity, shao2019long} among the 10 generations for each prompt (i.e., in total, 160,000 generations to evaluate the diversity of each method). 
\emph{iii)}~\textit{Repetition} is a common form of degeneration {that is very undesirable}. We measure the number of repetitive substrings that get generated at the end of the generations by using the publicly available metric code from~\citet{holtzman2019curious}.

\vspace{-1mm}
\subsection{Correlation with Human Judgement}
\vspace{-1mm}

\begin{wraptable}{r}{5.5cm}
\centering
\vspace{-4.5mm}
\small
\caption{\footnotesize{Pearson correlation coefficients between human factuality annotation and our factuality metrics. p-values for all results are $0.00$.}}
\vspace{-3mm}
\label{table:human_correlation}
\begin{tabular}{ccc}\\\toprule  
\textbf{Annotation} & \textbf{\entailedRatio} & \textbf{\halluNE} \\ \midrule
Expert & 0.81 & -0.77 \\  \midrule
Majority-voting & 0.47 & -0.46 \\
\bottomrule
\end{tabular}
\vspace{-2mm}
\end{wraptable} 



Although NE-based and entailment-based metrics have been used in downstream NLG tasks~\citep{ji2022survey}, they have not been utilized for evaluating factual accuracy in open-ended text generation. To ensure their validity, we collect human annotations to evaluate the correlation between our automatic factuality metrics with human judgement {-- i.e., are generations with higher \entailedRatio~and lower \halluNE~errors, more likely to be perceived as factual by human?}

We obtained human annotations for 200 randomly chosen LM continuations of varying \halluNE~and \entailedRatio~scores. The annotators are asked to fact-check the LM continuations against Wikipedia and assign factuality label (1 = Factual : can find supporting Wikipedia evidence. 0 = Non-factual : cannot find supporting Wikipedia evidence). 

The fact-checking annotation is a challenging and time-consuming task, as it requires the annotator to carefully read multiple evidences and reason over them. To improve the annotation quality, we have two types of annotations. The first type is two annotations from average English speaking workers on \emph{Appen.com} platform, and the second type is one ``expert'' annotation from one of the authors who is familiar with the task and  spent solid amount of time checking each samples. Based on these three annotations, we do majority voting and report the Pearson correlation results in Table~\ref{table:human_correlation}. We also report the correlation result solely using the expert annotations, and show that there is strong correlation between human judgement of factuality and the proposed automatic metric \halluNE~and \entailedRatio.
\halluNE~is negatively correlated with factuality because the lower the \halluNE~error, the better the factuality.


\begin{table}[]
\centering
\caption{\footnotesize
The factuality of LMs with different parameter size from 12M to 530B. \halluNE~refers to the named-entity error, \entailedRatio~refers to entailment ratio, Div. refers to distinct 4-grams, and Rep. refers to repetition. $\uparrow$ means the higher the better, and $\downarrow$ means the lower the better.}
\label{table:LM_size}
\vspace{1mm}
\small
\begin{tabular}{cc|cccc|cccc}
\toprule
\multirow{2}{*}{\textbf{Size}} & \multirow{2}{*}{\textbf{Decode}} & \multicolumn{4}{c}{\textbf{Factual Prompt}} & \multicolumn{4}{c}{\textbf{Nonfactual Prompt}} \\
\cmidrule(lr){3-6} \cmidrule(lr){7-10}
 &  & \halluNE$\downarrow$ & \entailedRatio$\uparrow$ & Div.$\uparrow$ & Rep.$\downarrow$ & \halluNE$\downarrow$ & \entailedRatio$\uparrow$ & Div.$\uparrow$ & Rep.$\downarrow$ \\ \midrule
\multirow{2}{*}{126M} & p=0.9 & 63.69\% & 0.94\% & 0.90 & 0.58\% & 67.71\% & 0.76\% & 0.90 & 0.38\% \\
 & greedy & 48.55\% & 8.36\% & 0.03 & 59.06\% & 54.24\% & 6.25\% & 0.03 & 59.90\% \\ \midrule
\multirow{2}{*}{357M} & p=0.9 & 56.70\% & 2.01\% & 0.87 & 0.55\% & 60.80\% & 1.42\% & 0.88 & 0.35\% \\
 & greedy & 43.04\% & 14.25\% & 0.03 & 45.18\% & 46.79\% & 9.89\% & 0.04 & 46.30\% \\ \midrule
\multirow{2}{*}{1.3B} & p=0.9 & 52.42\% & 2.93\% & 0.88 & 0.24\% & 56.82\% & 2.04\% & 0.89 & 0.25\% \\
 & greedy & 39.87\% & 12.91\% & 0.05 & 33.13\% & 45.02\% & 8.75\% & 0.05 & 36.20\% \\ \midrule
\multirow{2}{*}{8.3B} & p=0.9 & 40.59\% & 7.07\% & 0.90 & 0.11\% & 47.49\% & 3.57\% & 0.91 & 0.08\% \\
 & greedy & 28.06\% & 22.80\% & 0.07 & 19.41\% & 32.29\% & 15.01\% & 0.07 & 13.26\% \\ \midrule
\multirow{2}{*}{530B} & p=0.9 & 33.30\% & 11.80\% & 0.90 & 0.13\% & 40.49\% & 7.25\% & 0.92 & 0.08\% \\
 & greedy & \textbf{20.85\%} & \textbf{31.94\%} & 0.08 & 15.88\% & 27.95\% & 19.91\% & 0.08 & 16.28\% \\ \bottomrule
\end{tabular}
\label{tab:lm_size}
\vspace{-1mm}
\end{table}

\vspace{-1mm}
\section{Factuality Analysis of Pretrained LMs}
\label{sec:case_study}
\vspace{-1mm}
In this section, we perform a factuality analysis of LMs from three aspects: \emph{i)} model size, \emph{ii)} prompt type and \emph{iii)} decoding algorithm. 

\vspace{-3mm}
\paragraph{Model Size}
Researchers have observed the trend of larger LMs outperforming smaller ones in various downstream tasks~\citep{devlin2018bert, brown2020language, raffel2019exploring}. 
However, contradicting to these general observations, recent studies suggest that more misconceptions tend to be generated from larger models~\citep{lin2021truthfulqa}, 
and zero-shot fact-checking performance tend to stagnate with LM scaling~\citep{rae2021scaling}.
We study the factuality of LMs with a range of parameter sizes (126M, 357M, 1.3B, 8.3B, 530B) to understand whether such surprising trend also applies to open-ended text generation.
Note that, all LMs are pretrained on the same corpus as in \citep{smith2022using}.
As shown in Table~\ref{tab:lm_size}, generation factuality does improve with the scaling of model size, e.g., \halluNE~drops from 63.99\% to 33.30\% when parameter size scales up from 126M to 530B.

\vspace{-3mm}
\paragraph{Prompt Type}
Prompts provided to the LM are known to significantly affect the quality and characteristics of LM continuations~\citep{gehman2020realtoxicityprompts, wang2022exploring, wallace2019universal}. We use our factual and nonfactual prompts to test the behavior of LMs. Results in Table~\ref{table:LM_size} show that both factual and nonfactual prompts can lead to nonfactual generations, although factual prompts always result in less nonfactual generations. Interestingly, the performance gap between factual and nonfactual prompts gets more prominent as the model size increases ($4\%$ to $7\%$ in \halluNE~as parameter size increases from 126M to 530B). This could be due to the larger LM can better understand the prompts and imitate the factual or nonfactual prompts in the continuations.

\vspace{-3mm}
\paragraph{Decoding Algorithm}
We investigate the choice of decoding algorithms and their impacts on the factuality of generations. In particular, we compare two representative decoding algorithms that are \emph{greedy decoding}~(i.e., maximize generation likelihood) and \emph{nucleus sampling}~\citep{holtzman2019curious}. 
Nucleus sampling algorithm~(a.k.a. top-$p$) samples only from the top subword candidates with total {cumulative} probability $p$. It is popular for open-ended text generation because it solves the degeneration problems of the greedy decoding algorithm~(e.g., repetition). 
However, the results in Table~\ref{table:LM_size} show that 
top-$p$ decoding underperforms greedy decoding in terms of factuality, although it obtains higher generation diversity and less repetition. 
This intuitively makes sense because top-$p$ can be seen as adding ``randomness'' to encourage diversity, which as a result, can lead to factual errors. It is important to understand that factuality of a sentence can be easily altered by one wrong choice of word. For example, ``Barack Obama was born in 1961'' will be nonfactual if ``1961'' is changed to ``1962''.
In the same sense, greedy decoding is more factual because its way of choosing the word with the highest probability minimizes randomness and maximizes the utilization of parametric knowledge of LM~\citep{petroni2019language, jiang2020can}. However, greedy decoding 
sacrifices generation diversity and quality.

\vspace{-3mm}
\paragraph{Error Types}
We conduct a qualitative analysis of the factual errors from greedy generation of 530B LM, to understand what are the remaining errors when the randomness from decoding choice is strictly restricted. The two notable error types were:

\vspace{-2mm}
\begin{itemize}[leftmargin=2em]
    \item \textbf{Named Entity Mix-up}: Mixing up similar types of the named entity. For example, LM generated ``\textit{The movie is based on the novel of the same name by \textit{Gayle Forman}.}'' about a film called ``\textit{The Best of Me}''.
    However, the correct author's name is ``Nicholas Sparks'', not ``Gayle Forman''. Note that Gayle Forman is also an American young adult fiction author who writes similar type of novels as Nicholas Sparks.
    \vspace{-1mm}
    \item \textbf{Fabricated Fact:} Fabricating some random facts. For example, ``\textit{Samuel Witwer's father is a Lutheran minister}.'' Note that, the pretraining corpus contains non-factual or fictional information, which can also contribute to such fabricated facts. 
\end{itemize}
Both error types can be viewed as wrong associations of entities that appear at different parts of the training corpus with similar context.
Such behavior is unsurprising because these LMs are uniformly trained with the next subword prediction objective instead of a fact-related objective.

\begin{table}[]
\caption{\footnotesize
\textbf{1.3B} LM results with different decoding algorithms. \halluNE refers to named-entity error, \entailedRatio refers to entailed class ratio, Div. refers to distinct 4-grams, and Rep. refers to repetition. $\uparrow$ means the higher, the better, and $\downarrow$ means the lower, the better.
For factual-nucleus sampling, $p$, $\lambda$ and $\omega$ are nucleus probability, decay factor, and decay lowerbounds respectively. 
See more results with different hyperparameters in Figure~\ref{fig:diversity_vs_halluNE} and \ref{fig:repetition_vs_halluNE}.}
\vspace{1mm}
\label{table:decoding_choices}
\small
\centering
\begin{tabular}{lcccc|cccc}
\toprule
\multirow{2}{*}{\textbf{Decoding}} & \multicolumn{4}{c}{\textbf{Factual Prompt}} & \multicolumn{4}{c}{\textbf{Nonfactual Prompt}} \\ \cmidrule(lr){2-5} \cmidrule(lr){6-9}
 & \halluNE$\downarrow$ & \entailedRatio$\uparrow$ & \distFour$\uparrow$ & Rep.$\downarrow$ & \halluNE$\downarrow$ & \entailedRatio$\uparrow$ & \distFour$\uparrow$  & Rep.$\downarrow$ \\ \midrule
\emph{Greedy} & 39.9\% & 12.9\% & 0.05 & 33.1\% & 45.0\% & 8.8\% & 0.05 & 36.2\%  \\ 
\textit{Top-p 0.9} & 52.4\% & 2.9\% & 0.88 & 0.2\% & 56.8\% & 2.0\% & 0.89 & 0.3\%  \\
\midrule
\multicolumn{9}{l}{\ ~$p$ ~|~ $\lambda$ \hspace{.9cm}  Top-$p$ ~+~ \decay} \\
\midrule
0.9 | 0.9 & 41.1\% & 10.8\% & 0.43 & 30.7\%  & 45.7\% & 6.8\% & 0.47 & 34.5\%  \\
0.9 | 0.5 & 39.9\% & 13.0\% & 0.08 & 33.1\%  & 44.9\% & 9.1\% & 0.09 & 35.9\%  \\
\midrule
\multicolumn{9}{l}{\ ~$p$ ~|~ $\lambda$  \hspace{.9cm} Top-$p$ ~+~ \decay ~+~ $p$-reset } \\ \midrule
0.9 | 0.9 & 41.5\% & 10.3\% & 0.52 & 10.3\%  & 45.4\% & 6.3\% & 0.57 & 9.1\%  \\
0.9 | 0.5 & 39.3\% & 12.8\% & 0.34 & 17.8\%  & 44.5\% & 8.4\% & 0.45 & 18.9\%  \\
\midrule
\multicolumn{9}{l}{\ ~$p$ ~|~ $\lambda$ ~|~ $\omega$ \hspace{.3cm} Top-$p$ ~+~ \decay ~+~ \reset ~+~ \lowerBound~ (\emph{factual-nucleus sampling}) } \\ \midrule
0.9 | 0.9 | 0.7 & 46.2\% & 5.0\% & 0.78 & 1.2\% & 52.2\% & 3.2\% & 0.80 & 0.5\%  \\
0.9 | 0.9 | 0.3 & 42.1\% & 10.1\% & 0.55 & 7.1\% & 46.5\% & 5.6\% & 0.59 & 6.4\%  \\
0.9 | 0.9 | 0.2 & 41.7\% & 9.9\% & 0.52 & 8.6\% &  45.6\% & 6.2\% & 0.56 & 7.6\%\\
0.9 | 0.5 | 0.3 & 41.0\% & 12.2\% & 0.47 & 13.0\%  & 46.0\% & 7.0\% & 0.51 & 12.7\% \\
0.9 | 0.5 | 0.2 & 39.3\% & 12.8\% & 0.38 & 16.1\%  & 45.2\% & 7.8\% & 0.42 & 16.9\% \\ 
\bottomrule
\end{tabular}
\vspace{-1mm}
\end{table}

\begin{figure*}[h!]
     \centering
     \begin{subfigure}[b]{0.49\textwidth}
         \centering
         \includegraphics[width=\textwidth]{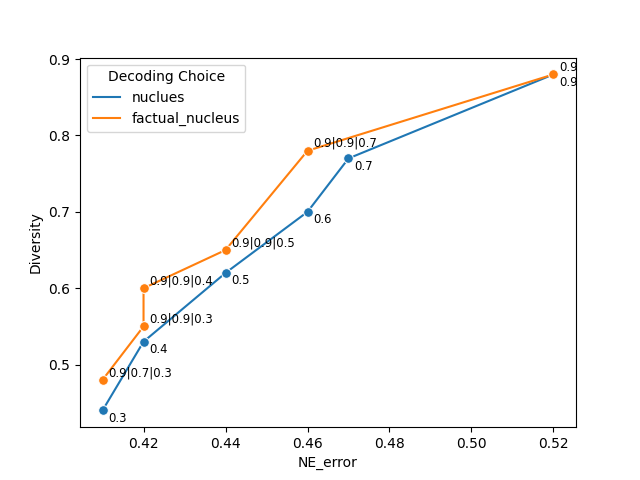}
         \caption{Diversity vs. \halluNE}
         \label{fig:diversity_vs_halluNE}
     \end{subfigure}
     \begin{subfigure}[b]{0.49\textwidth}
         \centering
         \includegraphics[width=\textwidth]{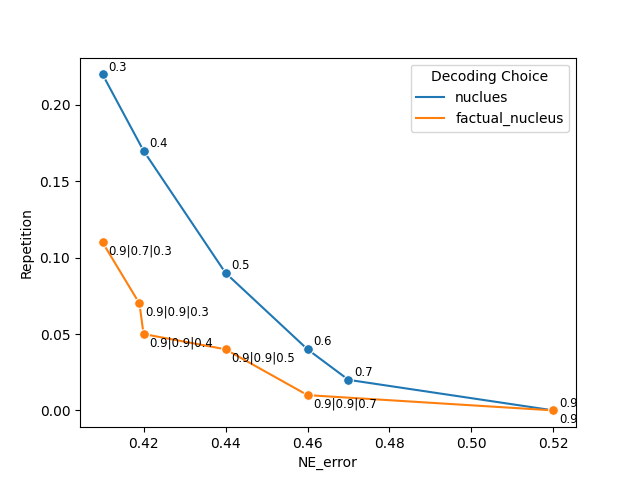}
         \caption{Repetition vs. \halluNE}
         \label{fig:repetition_vs_halluNE}
     \end{subfigure}
    \caption{Comparison between nucleus sampling (blue line)
    and factual-nucleus sampling (orange line).
    The x-axis is named entity error \halluNE. The y-axes are diversity and repetition in (a) and (b) respectively.
    The lower the repetition, the better. {It is evident that factual-nucleus sampling has better trade-offs between factuality and diversity/repetition.}
    For a reference, the diversity score of randomly sampled $5000$ Wikipedia documents is $0.767$.
    }
    \label{fig:topP_vs_ours}
\end{figure*}

\vspace{-1mm}
\section{Factual-Nucleus Sampling}
\label{sec:dynamic_topp}
\vspace{-1mm}
In this section, we propose a new sampling algorithm that achieves a better trade-off between generation quality and factuality than existing decoding algorithms.
\subsection{Method}
\vspace{-1mm}
We hypothesize that the randomness of sampling is more harmful to factuality when it is used to generate the latter part of a sentence than the beginning of a sentence. 
There is no preceding text at the start of a sentence, so it is safe for LM to generate anything as long as it is grammatical and contextual. However, as the generation proceeds, the premise become more determined, and fewer word choices can make the sentence factual. Given the example ``\textit{Samuel Witwer's father is a Lutheran minister}'', the beginning of the sentence ``\textit{Samuel Witwer's father is}'' is not nonfactual. However, the continuation of ``\textit{Lutheran minister}'' makes the sentence nonfactual. 
Therefore, we introduce the \emph{factual-nucleus sampling} algorithm that dynamically adapts the ``{nucleus}'' $p$ along the generation of each sentence.
In \emph{factual-nucleus sampling}, the nucleus probability $p_{t}$ to generate the $t$-th token within each sentence is,
$$
p_{t} = \max \{\omega,~  p \times \lambda^{t-1}\},
$$
where $\lambda$ is the decay factor for top-$p$ probability, and $\omega$ lower bounds the decay of probability. Specifically, it has the following parts:
\vspace{-2mm}
\begin{itemize}[leftmargin=2em]
    \item \textbf{\decay}: Given that top-$p$ sampling pool is selected as a set of subwords whose cumulative probability exceeds $p$, we gradually decay the $p$ value with decay factor $\lambda$ at each generation step to reduce the ``randomness'' through time.
    \vspace{-0.1em}
    \item \textbf{\reset}: 
    The nucleus probability $p$ can quickly decay to a small value after a long generation. So, we reset the $p$-value to the default value at the beginning of every new sentence in the generation (we identify the beginning of a new sentence by checking if the previous step has generated a full-stop). This reduces the unnecessary cost of diversity for any long generations. 
    \vspace{-0.1em}
    \item \textbf{\lowerBound}:
     If \decay~is applied alone, the $p$-value could become too small to be equivalent to greedy decoding and hurt diversity. To overcome this, we introduce a lower-bound $\omega$ to limit how far $p$-value can be decayed.
\vspace{-0.4em}
\end{itemize}
We will show the importance of each parts with ablation studies.

\vspace{-1mm}
\subsection{Result}
\vspace{-1mm}
We report our decoding experimental results with 1.3B LM~\footnote{1.3B LM is mainly used as it is big enough to have good learning capacity yet not too resource expensive.} in Table~\ref{table:decoding_choices}. 
Additions of \decay~helps improve top-$p$ 0.9 factuality results -- for instance, with decay rate $\lambda$ = 0.5, there is 12.5\% drop in \halluNE~and 10.1\% gain in \entailedRatio. However, this affects the diversity and repetition to become similar to greedy decoding.
\reset~mitigates the repetition issue and improves diversity metric without losing much in factuality metric. The effect is more drastic for the $\lambda$ = 0.5 option, where it achieves 0.26 gains in diversity metric with negligible changes in factuality scores.  
By also adding \lowerBound, we obtain the anticipated factuality performance (i.e., similar to greedy decoding), with great improvement in generation quality over greedy; with $p$=0.9, $\lambda$=0.9, $\omega$=0.3, we achieve $\times$11 improvement in diversity and $\times$4.6 improvement in repetition over greedy. 
Although our factual-nucleus sampling still under-performs top-$p$ 0.9 in terms of diversity, we believe this is an acceptable trade-off to improve the factuality of LM for factually sensitive open-ended generation tasks. 
Our proposed decoding does not harm the sentence fluency; its perplexity do not exceed the perplexity of top-p. {Refer to Appendix~\ref{appendix:full_table_with_ppl} for full perplexity results}.

To further illustrate the underlying trade-off, we also compare the proposed factual-nucleus sampling against the nucleus sampling with lower $p$ values that are also expected to have lower randomness, thus less factual error, in generations.
Specifically, we plotted results for nucleus sampling with $p$ = $\{0.9, 0.7, 0.6, 0.5, 0.4, 0.3\}$, and factual nucleus sampling with the following $p~|~\lambda~|~\omega$ choices: {0.9|0.9|0.7, 0.9|0.9|0.5, 0.9|0.9|0.4, 0.9|0.9|0.3, 0.9|0.7|0.3.}
The Fig~\ref{fig:diversity_vs_halluNE} and Fig~\ref{fig:repetition_vs_halluNE} respectively show that the factual nucleus sampling method has better trade-offs than top-$p$ in factuality-vs-diversity and factuality-vs-repetition. 
In other words,  it always achieves better factuality score with the same level of diversity and repetition scores.

\vspace{-1mm}
\section{Factuality-Enhanced Continued Training}
\label{sec:factual_training}
\vspace{-1mm}
This section introduces factuality-enhanced method for continued training of LMs~\citep{gururangan2020don}. We introduce the {\prefix} for better awareness of facts and the sentence completion loss as training objective.

\vspace{-1mm}
\subsection{Prepending {\prefix}}
\vspace{-1mm}
Unstructured factual knowledge typically exists at a document level (i.e., a group of factual sentences about an entity). This means that sentences can contain pronouns (e.g., she, he, it), making these sentences factually useless standalone.
To illustrate with an example from Barack Obama's Wikipedia page, ``He previously served as a U.S. senator from Illinois from 2005 to 2008'' cannot be a useful standalone fact because it is unclear who ``He'' is. 
Due to the GPU memory limit and computation efficiency, it is common to chunk documents in LM training corpus. This causes the ``fragmentation'' of information and leads to wrong associations of entities that appear in independent documents with similar contexts. As a remedy, we propose to prepend {\prefix} to sentences in the factual documents to make each sentence serve as a standalone fact. In our experiments, we mainly utilize Wikipedia as the factual corpus and the Wikipedia document name as the {\prefix}.

\vspace{-1mm}
\subsection{Sentence Completion Loss}
\vspace{-1mm}
We propose a sentence completion loss to address the incorrect association learned between entities. 
To explain our rationale, let us recall the nonfactual example from \S\ref{sec:dynamic_topp}: ``\textit{Samuel Witwer's father is a Lutheran minister}''. This sentence is nonfactual because LM failed to generate factually correct information after ``\textit{is}''. 
In other words, LM failed to accurately \textit{complete} the sentence given the generated context. One reason is that the LM is uniformly trained to predict each subword token within the sentence, when ensuring the correct prediction at the latter section of sentence is more critical for factuality.
Therefore, we construct a sentence completion loss, which makes the LM focus on predicting the subwords later in the sentence. 
For implementation, we determine a pivot $t$ for each sentence, and then apply zero-masking for all token prediction losses before $t$. \yeon{This pivot is only required during training (i.e., no pivot needed during inference time).}

We emphasize that this loss masking is different from the input token masking applied in BERT~\citep{devlin2018bert} or BART~\citep{,lewis2019bart}, and the LM is still trained in an autoregressive manner.
Note that many BART-based summarization models are known to still suffer from factual errors, suggesting that masked prediction at the encoder level may not effectively transfer well to autoregressive text generation. 

In this work, we explore three strategies (from simple to complex) to determine the pivot $t$:
\begin{itemize}[leftmargin=2em]
\vspace{-2mm}
    \item \halfSC: pivot $t = 0.5 \times$ sentence-length.
    \vspace{-1mm}
    \item \randomSC: random pivot, e.g., $t \sim \text{uniform}[0.25, 0.75] \times$ sentence-length.
    \vspace{-1mm}
    \item \rootSC: pivot $t =$ position of ROOT (relation) from dependency parsing.
\vspace{-2mm}
\end{itemize}
\yeon{Our experiments show that the simplest \halfSC~performs on par with the complex ones (such as {\rootSC}), thus, we suggest future work to choose \halfSC~strategy.}

\begin{table}[]
\small
\centering
\caption{\footnotesize
Results for factuality enhanced training. The decoding settings are formatted as: nucleus probability $p$, decay rate $\lambda$, lower-bound $\omega$. }
\small
\vspace{1mm}
\label{table:factuality_enhanced_training}
\begin{tabular}{lcccc|cccc}
\toprule
\multirow{2}{*}{\textbf{Decoding}} & \multicolumn{4}{c}{\textbf{Factual Prompt}} & \multicolumn{4}{c}{\textbf{Nonfactual Prompt}} \\
\cmidrule(lr){2-5} \cmidrule(lr){6-9}
\textbf{($p$ | $\lambda$ | $\omega$)} & \multicolumn{1}{c}{\halluNE$\downarrow$} & \multicolumn{1}{c}{\entailedRatio$\uparrow$} & \multicolumn{1}{c}{\distFour} & \multicolumn{1}{c}{Rep.} & \multicolumn{1}{c}{\halluNE} & \entailedRatio & \distFour & Rep. \\ \midrule
\multicolumn{9}{l}{Vanilla Pretrained LM (1.3B) 
}\\ \midrule
0.9 & 52.4\% & 2.9\% & 0.88 & 0.2\% & 56.8\% & 2.0\% & 0.89 & 0.3\% \\
0.9 | 0.9 | 0.3 & 42.1\% & 10.1\% & 0.55 & 7.1\% & 46.5\% & 5.6\% & 0.59 & 6.4\%  \\
\midrule
\multicolumn{9}{l}{Factual Domain-Adaptive Training with Wikipedia (1.3B)}\\ \midrule
0.9 & 52.5\% & 2.8\% & 0.85 & 0.2\%  & 55.8\% & 2.2\% & 0.86 & 0.1\% \\
0.9 | 0.9 | 0.3 & 42.7\% & 7.1\% & 0.51 & 7.2\%  & 48.2\% & 4.9\% & 0.56 & 6.0\%\\ \midrule
\multicolumn{9}{l}{{\prefix} (1.3B)} \\ \midrule
0.9 & 34.4\% & 4.2\% & 0.84 & 0.3\% & 36.2\% & 2.7\% & 0.85 & 0.2\% \\
0.9 | 0.9 | 0.3 & 27.6\% & 8.7\% & 0.43 & 8.0\% & 30.5\% & 6.1\% & 0.47 & 6.9\% \\ \midrule
\multicolumn{9}{l}{{\prefix} + \rootSC~(1.3B)}  \\ \midrule
0.9 & 32.5\% & 6.7\% & 0.83 & 1.2\% & 34.3\% & 4.6\% & 0.84 & 1.1\% \\
0.9 | 0.9 | 0.3 & 24.7\% & 15.8\% & 0.40 & 13.6\%  & 27.6\% & 9.1\% & 0.44 & 13.7\% \\ \midrule
\multicolumn{9}{l}{{\prefix} + \randomSC~(1.3B)} \\ \midrule
0.9 & 32.0\% & 7.9\% & 0.81 & 1.2\%  & 34.2\% & 5.5\% & 0.83 & 1.1\% \\
0.9 | 0.9 | 0.3 & 23.6\% & 17.6\% & 0.39 & 14.2\%  & 26.9\% & 9.3\% & 0.42 & 13.2\%  \\ \midrule
\multicolumn{9}{l}{{\prefix} + \halfSC~(1.3B)}  \\ \midrule
0.9 & 31.6\% & 7.6\% & 0.81 & 1.4\%  & 33.5\% & 5.1\% & 0.83 & 1.5\%  \\
0.9 | 0.9 | 0.3 & 23.6\% & 17.4\% & 0.38 & 14.4\%  & 27.2\% & 10.2\% & 0.42 & 13.1\%  \\ \midrule
\multicolumn{9}{l}{Vanilla Pretrained LM (530B)} \\ \midrule
0.9 & 33.3\% & 11.8\% & 0.90 & 0.1\% & 40.5\% & 7.25\% & 0.92 & 0.1\% \\
\midrule
\multicolumn{9}{l}{{\prefix} + \halfSC~(530B)} \\ \midrule
0.9 & 18.3\% & 19.3\% & 0.68 & 0.1\%  & 21.7\% & 13.7\% & 0.68 & 0.1\%  \\
0.9 | 0.9 | 0.3 & \textbf{14.5\%} & \textbf{25.5\%} & 0.33 & 0.2\% & \textbf{17.7\%} & \textbf{20.0\%} & 0.33 & 0.1\%  \\
\bottomrule
\end{tabular}
\vspace{-1mm}
\end{table}

\vspace{-1mm}
\subsection{Results}
\vspace{-1mm}
The results are reported in Table~\ref{table:factuality_enhanced_training}, and experimental setups are reported in Appendix~\ref{appendix:experiment_setup}.

\textbf{Inefficiency of Domain Adaptive Training}~\  
The pre-training corpus of LM contains both factual texts (e.g., Wikipedia) and potentially nonfactual texts (e.g., rumors, fake news)~\footnote{See \citep{smith2022using} for details of pre-training corpus.}. The nonfactual domain of the training corpus could be the problem. Thus, we conduct a baseline experiment that does domain-adaptive training with strictly factual domain text only (i.e., Wikipedia). Interestingly, we find that domain-adaptive training can hardly improve generation factuality.

\textbf{Effect of {\prefix}}~\  
Continued pre-training of 1.3B LM with {\prefix} preprocessed Wikipedia alone can already improve the factuality, especially in terms of \halluNE.  
For example, it reduces the \halluNE from $42.1\%$ to $27.6\%$ when we use the factual-nucleus decoding~(0.9 | 0.9 | 0.3), which even outperforms the 1.3B with greedy decoding (\halluNE: $27.6\%$ vs. $39.9\%$) with much less repetition~($8.0\%$ vs. $33.1\%$). 

\textbf{Effect of Sentence Completion Loss}~\   
The proposed sentence completion loss further helps to improve the factuality, especially for the \entailedRatio. For example, when one uses factual-nucleus decoding on trained 1.3B model, {\prefix} + \halfSC~can further improve \entailedRatio~from $8.7\%$ to $17.4\%$ than {\prefix} alone, while reducing \halluNE~from $27.6\%$ to $23.6\%$.
Note that the results show consistent improvement across different pivot selection strategies, suggesting that the sentence completion loss is robust.
In particular, the simplest \halfSC~ performs as good as others or even better in terms of several metrics. Thus we recommend it as the default option.

\textbf{530B vs 1.3B}~\ 
As expected, our method on 530B LM further reduces the factual errors and achieves the lowest \halluNE~($14.5\%$) and the highest \entailedRatio~($25.5\%$). 
Surprisingly, our method on 530B LM lead to less diverse generation than 1.3B LM despite the significant improvement in the generation quality (i.e., near perfect repetition scores $0.1\% ~ 0.2\%$). 
We conjecture that this is the trade-off between the factuality and diversity for 530B LM.


\vspace{-1mm}
\section{Conclusion}
\label{sec:conclusion}
\vspace{-1mm}
In this work, we establish a benchmark to measure and analyze factuality in open-ended text generation tasks. We propose \emph{factual-nucleus sampling} that improves generation factuality at inference time, and the combination of sentence completion loss and {\prefix} pre-processing that improves factuality with continued training. 
We demonstrate that our methods are effective in improving the factuality. 
Lastly, our results shed light on the existence of 
the trade-off between diversity and factuality. We strongly believe this is an important insight that will help researchers make a better-informed decision about their model design - i.e., appropriately prioritize the desirable attribute of their LM (factuality vs. diversity) according to the final goal of their task. Potential future work would be to reduce the degree of the observed trade-offs. 




{
\small
\bibliography{main}
\bibliographystyle{unsrtnat}
}

\newpage

\newpage

\appendix


\vspace{-2mm}
\section{Generation Examples}
\label{appendix:more_samples}
\vspace{-2mm}
We provide more generation examples from the pretrained 530B LM with greedy and top-$p$ sampling ($p=0.9$), and factuality-enhanced 530B LM with factual-nucleus sampling~(Ours).
\textcolor{aogreen}{Green} indicates factual, \textcolor{red}{red} indicates nonfactual, and \st{striked text} indicates repetition. 
Refer to Appendix~\ref{appendix:more_generations_2} for more examples. Disclaimer: Authors tried to exhaustively check the factuality of the following generations, however, there is no 100\% guarantee about the annotations.

\begin{table}[h!]
\small
\vspace{-1mm}
\centering
\begin{tabular}{p{0.95\linewidth}}
\toprule
\textbf{[Category]} Medical \\
\textbf{[Factual Prompt]} One of lung cancer's symptoms is shortness of breath. \\ \midrule
\textbf{[Greedy]} \textcolor{aogreen}{The American Cancer Society estimates that more than 228,000 people will be diagnosed with lung cancer in the United States this year.\textbackslash{}n\textbackslash{}n}
\textcolor{aogreen}{
\st{The American Cancer Society estimates that more than 228,000 people will be diagnosed with lung cancer in the United States this year.\textbackslash{}n\textbackslash{}nThe American Cancer Society estimates that more than 228,000 people will be diagnosed with lung cancer in the United...}} \\
\vspace{0mm}
\textbf{[p=0.9]} \textcolor{red}{That would make an oxygen mask one of the more popular treatments for this devastating disease.} \textcolor{aogreen}{It helps ease breathing and give patients back their strength.} Learn more here.<|endoftext|>" \\
\vspace{0mm}
\vspace{0mm}
\textbf{[Ours]} \textcolor{aogreen}{Other symptoms include coughing (including coughing up blood), chest pain, \textcolor{red}{hoarseness}, weight loss, and loss of appetite. The symptoms of lung cancer may resemble those of a bad cold or pneumonia. A chest x - ray is \textcolor{red}{necessary} to confirm the diagnosis. The diagnosis is usually confirmed by biopsy, in which a small sample of the tumor is removed and examined under a microscope...} 
\\ 
\midrule
\textbf{[Category]} Actor \\
\textbf{[Factual Prompt]} Leonardo DiCaprio played a supporting role in What's Eating Gilbert Grape. \\ \midrule
\textbf{[Greedy]} \textcolor{red}{The film was shot in the town of Endora, Iowa.\textbackslash{}n\textbackslash{}n}\textcolor{red}{\st{The film was shot in the town of Endora, Iowa.\textbackslash{}n\textbackslash{}nThe film was shot in the town of Endora, Iowa.\textbackslash{}n\textbackslash{}nThe film was shot in the town of Endora, Iowa.\textbackslash{}n\textbackslash{}nThe film was shot in the town of Endora, Iowa.\textbackslash{}n\textbackslash{}nThe film was shot in the town of Endora, Iowa.\textbackslash{}n\textbackslash{}nThe film was shot in the town of Endora, Iowa.\textbackslash{}n\textbackslash{}nThe film was shot in the town of Endora, ...}} \\
\vspace{0mm}
\textbf{[p=0.9]} \textcolor{red}{DiCaprio was desperate to have the lead role in the film. However, after realizing the amount of time that was required to learn all the lines, he gave the role to Johnny Depp instead.\textbackslash{}n\textbackslash{}nTitanic (1997)\textbackslash{}n\textbackslash{}nLeonardo DiCaprio and Kate Winslet were paid the same amount for their roles in Titanic.\textbackslash{}n\textbackslash{}nAlthough DiCaprio was relatively unknown when the film was released, Paramount Pictures paid him and Winslet the same amount.} \\
\vspace{0mm}
\textbf{[Ours]} \textcolor{aogreen}{In 1993 , he played the mentally handicapped brother of Johnny Depp 's character in " What's Eating Gilbert Grape" , a comic - tragic odyssey of a dysfunctional Iowa family. Director Lasse Hallström admitted he was initially looking for a less good - looking actor but finally settled on DiCaprio as he had emerged as "the most observant actor" among all who auditioned . Budgeted at US\$ 11 million , the film became a critical success, resulting in various accolades for DiCaprio , who was awarded the National Board of Review Award and nominated for both an Academy Award and a Golden Globe Award for his portrayal of Arnie Grape...} \\
\midrule \midrule 
\textbf{[Category]} Company \\
\textbf{[Factual Prompt]} KFC is the second biggest restaurant chain in the world. \\ \midrule
\textbf{[Greedy]} \textcolor{aogreen}{The company is owned by Yum! Brands, which also owns Pizza Hut and Taco Bell.\textbackslash{}n\textbackslash{}nKFC is known for its fried chicken, but it also offers other food items like chicken sandwiches, wraps, salads, and even rice bowls.\textbackslash{}n\textbackslash{}n\st{KFC is known for its fried chicken, but it also offers other food items like chicken sandwiches, wraps, salads, and even rice bowls.\textbackslash{}n\textbackslash{}nKFC is known for its fried chicken, but it also offers other food items like chicken sandwiches, wraps, salads, and even rice bowls.\textbackslash{}n\textbackslash{}nKFC...}}
\\
\vspace{0mm}
\textbf{[p=0.9]} \textcolor{red}{While many restaurants worldwide have been growing their mobile commerce revenues, KFC chose to implement its payment system via cash. Customers must visit a traditional cash register to buy food.\textbackslash{}n\textbackslash{}nThe order-and-pay system that KFC had been using at its stores didn't allow customers to pay with their smartphones. Instead, customers have to stand in long lines to order and pay for their meals. The process involves taking a physical receipt, which is then placed at the counter, to process the order.\textbackslash{}n\textbackslash{}n...}
\\
\vspace{0mm}
\textbf{[Ours]} \textcolor{aogreen}{KFC was founded by Colonel Harland Sanders , an entrepreneur who began selling fried chicken from his roadside restaurant in Corbin , Kentucky , during the Great Depression . Sanders identified the potential of the restaurant franchising concept , and the first " Kentucky Fried Chicken " franchise opened in Utah in 1952 . KFC popularized chicken in the fast food industry , diversifying the market by challenging the established dominance of the hamburger . By branding himself as " Colonel Sanders " , Harland became a prominent figure of American cultural history , and his image remains widely used in KFC advertising to this day...}
\\ \bottomrule
\end{tabular}
\end{table}

\begin{table}[h!]
\small
\centering
\begin{tabular}{p{0.95\linewidth}}
\toprule
\textbf{[Category]} History \\
\textbf{[Factual Prompt]} San Francisco was first colonized by Spanish colonists. \\ \midrule
\textbf{[Greedy]} \textcolor{red}{The Spanish arrived in 1776} and \textcolor{aogreen}{established the Presidio of San Francisco in 1776. The Mission San Francisco de Asís was founded in 1776. The Mission was the first European settlement in the city. \st{The Mission was the first European settlement in the city. The Mission was the first European settlement in the city. The Mission was the first European settlement in the city. The Mission was the first European settlement in the city. The Mission was the first European settlement in the city.}}
\\
\vspace{0mm}
\textbf{[p=0.9]}
\textcolor{aogreen}{A Spaniard, José Joaquin Moraga, who accompanied the De Anza expedition}, \textcolor{red}{named the bay, which he called "Puerto de San Francisco", on or about October 30, 1776. On January 12, 1777, the Presidio of San Francisco, founded by Moraga, was established by the Spanish. The Mission San Francisco de As\u00eds, established in the same year, was the first Mission on the Pacific Coast of North America. In 1781, the settlers, numbering about 200...}
\\
\vspace{0mm}
\textbf{[Ours]} \textcolor{aogreen}{The Presidio of San Francisco , a Spanish military outpost , was established in 1776 . It was the first European settlement in what is now the city . The Mission San Francisco de Asís ( Mission Dolores ) , established in 1776 , was the city 's first major settlement for the next half - century . The California Gold Rush of 1849 brought rapid growth , making it the largest city on the West Coast at the time . San Francisco became a consolidated city - county in 1856 . San Francisco 's status as the West Coast 's largest city peaked between 1870 and 1900 , when around 25 \% of California 's population resided in the city proper . After three - quarters of the city was destroyed by the 1906 earthquake and fire , San Francisco was quickly rebuilt...}
\\
\midrule \midrule

\textbf{[Category]} Sports \\
\textbf{[Factual Prompt]} Michael Jordan helped popularize the NBA in the 80s and 90s. \\ \midrule
\textbf{[Greedy]} \textcolor{aogreen}{The NBA is the most popular basketball league in the world.\textbackslash{}n\textbackslash{}n\st{The NBA is the most popular basketball league in the world.\textbackslash{}n\textbackslash{}nThe NBA is the most popular basketball league in the world.\textbackslash{}n\textbackslash{}nThe NBA is the most popular basketball league in the world.\textbackslash{}n\textbackslash{}nThe NBA is the most popular basketball league in the world.\textbackslash{}n\textbackslash{}nThe NBA is the most popular basketball league in the world.\textbackslash{}n\textbackslash{}nThe NBA is the most...}}
\\
\vspace{0mm}
\textbf{[p=0.9]} \textcolor{aogreen}{He was named the most valuable player in the NBA Finals six times. But the former Chicago Bulls star and current Charlotte Hornets owner} \textcolor{red}{has another MVP title to add to his list: MVP of Valentine's Day. An online sports book claims that Jordan is the most searched player of all time on Feb. 14.\textbackslash{}n\textbackslash{}n} >> Read more trending news\textbackslash{}n\textbackslash{}n\textcolor{red}{FanDuel Sportsbook, an online gaming company, said it reviewed searches on its site for the last seven years and discovered that Jordan is the most searched-for NBA player on Valentine's Day, Forbes",}
\\
\vspace{0mm}
\textbf{[Ours]} \textcolor{aogreen}{Jordan 's individual accolades and accomplishments include six NBA Finals Most Valuable Player ( MVP ) Awards , ten scoring titles ( both all - time records ) , five MVP Awards , ten All - NBA First Team designations , nine All - Defensive First Team honors , fourteen NBA All - Star Game selections , three All - Star Game MVP Awards , three steals titles , and the 1988 NBA Defensive Player of the Year Award. He holds the NBA records for highest career regular season scoring average ( 30.12 points per game ) and highest career playoff scoring average ( 33.45 points per game ). Jordan is also known for his product endorsements . He fueled the success of Nike 's Air Jordan sneakers , which were introduced in 1984...}
\\ \midrule \midrule

\textbf{[Category]} TV \\
\textbf{[Factual Prompt]} In Too Deep is an American crime thriller film. \\ \midrule
\textbf{[Greedy]} \textcolor{aogreen}{It was directed by Michael Rymer and released in 1999. The film stars Omar Epps, LL Cool J, Nia Long, Stanley Tucci}, \textcolor{red}{and Giancarlo Esposito. A young man named Michael (Omar Epps) is released from prison after serving a sentence for drug dealing. He is determined to go straight, but his friend Tommy (LL Cool J) is not so sure. Tommy is a drug dealer who has been working for} \textcolor{aogreen}{a crime boss named Dwayne Gittens} \textcolor{red}{(Stanley Tucci)}. \textcolor{aogreen}{Gittens is a violent man who will not hesitate to kill anyone who crosses him.\textbackslash{}n\textbackslash{}n} \textcolor{red}{Michael is soon dragged back into the drug world when Tommy is arrested for killing a man who tried to rob him. Gittens tells Michael that he will free Tommy if Michael will sell a large amount of drugs...} \\
\vspace{0mm}
\textbf{[0.9]} \textcolor{red}{The plot concerns the federal government's attempts to lure a bank robber out of hiding by teaming up a convicted high-profile criminal with a U.S. Marshal. \textcolor{aogreen}{It stars Omar Epps and LL Cool J and is directed by Michael Rymer.}\textbackslash{}n\textbackslash{}nThe film is also known as Deeper and In Too Deep. The film had a DVD release on December 26, 2000, and an HD DVD release on April 30, 2007.\textbackslash{}n\textbackslash{}nPlot\textbackslash{}nRonnie Pullman (Omar Epps) is a thief who has made a name for himself by stealing from...} \\
\vspace{0mm}
\textbf{[Ours]}
\textcolor{aogreen}{
It was directed by Michael Rymer from a screenplay written by Michael Henry Brown and Paul Aaron . The film stars Omar Epps , LL Cool J , Stanley Tucci , Pam Grier and Nia Long. Jeffrey Cole (Omar Epps) is a recent graduate of the Cincinnati police academy who dreams of working undercover. 
Cole manages to get an undercover assignment the day of graduation and earns the praise of his superiors . He is soon given the task of taking down statewide crack dealer Dwayne Gittens ( LL Cool J ) , an underworld boss so powerful that his nickname is "God"} 
\\
\bottomrule
\end{tabular}
\end{table}

\clearpage
\begin{table}[h!]
\small
\centering
\begin{tabular}{p{0.95\linewidth}}
\toprule
\textbf{[Category]} Medical \\
\textbf{[Non-Factual Prompt]} Obesity and endometrial cancer have nothing to do with each other. \\ \midrule
\textbf{[Greedy]} \textcolor{aogreen}{\textbackslash{}n\textbackslash{}nObesity is a risk factor for endometrial cancer. \textbackslash{}n\textbackslash{}n}\textcolor{red}{Obesity is not a risk factor for endometrial cancer.\textbackslash{}n\textbackslash{}n}Correct\textbackslash{}n\textbackslash{}nIncorrect\textbackslash{}n\textbackslash{}nQuestion 3 of 10\textbackslash{}n\textbackslash{}n3. Question\textbackslash{}n\textbackslash{}nWhich of the following statements is true?\textbackslash{}n\textbackslash{}n\textcolor{aogreen}{Endometrial cancer is the most common gynecologic cancer in the United States.\st{\textbackslash{}n\textbackslash{}nEndometrial cancer is the most common gynecologic cancer in the United States.\textbackslash{}n\textbackslash{}nEndometrial cancer is the most common gynecologic cancer in the United States.\textbackslash{}n\textbackslash{}nEndometrial cancer is the most common gynecologic cancer in the United States.}}
\\
\vspace{0mm}
\textbf{[p=0.9]} \textcolor{red}{\textbackslash{}n\textbackslash{}nThen why does it get such a high incidence rate with obese women?\textbackslash{}n\textbackslash{}nIt doesn't. The problem is that fat women are more likely to have abnormal uterine bleeding.\textbackslash{}n\textbackslash{}nAnd with an enlarged uterus, it may be difficult for a physician to detect a...} \\
\vspace{0mm}
\textbf{[Ours]} \textcolor{aogreen}{Obesity increases the risk of endometrial cancer . This is thought to be due to higher levels of estrogens produced by the adipose tissue . Obesity also causes changes in the endometrium that make it more susceptible to cancer...} \\

\midrule \midrule

\textbf{[Category]} Sports \\
\textbf{[Non-Factual Prompt]} Michael Jordan was inactive in the 1980s and 90s. \\ \midrule
\textbf{[Greedy]}
\textcolor{red}{The Chicago Bulls were inactive in the 1980s and 90s.\textbackslash{}n\textbackslash{}n\st{The Chicago Bulls were inactive in the 1980s and 90s.\textbackslash{}n\textbackslash{}nThe Chicago Bulls were inactive in the 1980s and 90s.\textbackslash{}n\textbackslash{}nThe Chicago Bulls were inactive in the 1980s and 90s...}}
\\
\vspace{0mm}
\textbf{[p=0.9]} \textcolor{red}{This caused concern that the new coach was not doing anything right. But there was still hope because he promised the following: \"We're going to be active traders on the phones\". Except it turned out he really meant they were going to be doing a lot of long distance calling and give...}
\\ \vspace{0mm}
\vspace{0mm}
\textbf{[Ours]} \textcolor{aogreen}{Jordan 's first retirement in 1993 was very short - lived , as he returned to the NBA in March 1995 , wearing the number 45 , after his father was murdered .\textbackslash{}n The Bulls were then eliminated in the second round of the playoffs by the \textcolor{red}{Orlando Magic.\textbackslash{}n}} \textcolor{red}{Jordan was suspended by the NBA for...}
\\
\midrule \midrule
\textbf{[Category]} TV \\
\textbf{[Non-Factual Prompt]} Modern Family is solely a Dutch television sitcom. \\ \midrule
\textbf{[Greedy]} \textcolor{red}{It is the first Dutch television series to be produced in HD. The series is produced by the Dutch broadcasting company Talpa, which had previously produced the successful series Tienstok. The series is written by Dutch screenwriter and director Joram Lürsen.\textbackslash{}n\textbackslash{}nThe series is about a...}
\\
\vspace{0mm}
\textbf{[p=0.9]} \textcolor{red}{It follows the story of a quite stereotypical Dutch family (although unlike the Dutch, the family in this sitcom has only two kids and only the youngest is still a minor, whereas in reality most Dutch families have at least three kids). The majority of the episodes are just ordinary family life, but the writers do not...}
\\
\vspace{0mm}
\textbf{[Ours]} \textcolor{red}{The show premiered on Dutch television on 18 September 2007 on RTL 4 .} \textcolor{aogreen}{The show was originally created by Steve Levitan and Christopher Lloyd , who also serve as executive producers . The series is set in the \textcolor{red}{fictional suburban town of Dunphy}, California . The show follows the lives of the Dunphy family , consisting of Phil Dunphy \textcolor{red}{( Eric Stonestreet )} , Claire Dunphy ( Julie Bowen ) , \textcolor{red}{Cameron Dunphy} ( Jesse Tyler Ferguson ) , Mitchell \textcolor{red}{Dunphy ( Ty Burrell )} , and Lily \textcolor{red}{Dunphy ( Sofia Vergara )...}}
\\ 
\midrule \midrule
\textbf{[Category]} Singer \\
\textbf{[Non-Factual Prompt]} Michael Jackson made his debut with his sister Janet. \\ \midrule
\textbf{[Greedy]} \textcolor{red}{\textbackslash{}n\textbackslash{}nThe Jackson 5 made their debut on the show in 1969.\st{\textbackslash{}n\textbackslash{}nThe Jackson 5 made their debut on the show in 1969.\textbackslash{}n\textbackslash{}nThe Jackson 5 made their debut on the show in 1969.\textbackslash{}n\textbackslash{}nThe Jackson 5 made their debut on the show in 1969.\textbackslash{}n\textbackslash{}nThe Jackson 5...}}
\\
\vspace{0mm}
\textbf{[p=0.9]} \textcolor{red}{Two performances followed before Jackson would begin a solo career, while \textcolor{aogreen}{still continuing to perform with his brothers in The Jacksons.}\textbackslash{}n\textbackslash{}n1977\textbackslash{}n\textbackslash{}nMichael Jackson releases his first solo album 'Off the Wall'. A \textcolor{aogreen}{critical and commercial success}, 'Off the Wall' reached No.5 in the UK album...}
\\
\vspace{0mm}
\textbf{[Ours]} \textcolor{aogreen}{The Jackson 5 won several awards within the \textcolor{red}{black music community , including the American Music Award for Favorite Soul / R\&B Band / Duo} in 1970 and 1972 . Jackson 's solo career began in 1971 while at Motown with his first solo album , " Got to Be There " , which was certified platinum . His second album , " Ben " , was released in 1972 . The album sold over a million copies , and the single " Ben " reached number one on the " Billboard " Hot 100 . Jackson 's third album , " Music \& Me " , was released in 1973 . In 1975 , Jackson moved to Epic Records...}
\\
\bottomrule
\end{tabular}
\end{table}

\clearpage
\section{Details about Claim Filtering Step in \S\ref{sec:benchmark}}
\label{appendix:claim_filtering}
The goal of open-ended text generations does not require all generations to always contain ``facts''. There can be generations that are perfectly grammatical and fluent, yet do not contain any checkworthy content such as personal opinions and daily small talks. 
Thus, we filter out ``not-checkworthy'' sentences that possess any of the following characteristics:

\vspace{-2mm}
\begin{itemize}[leftmargin=2em]
    \item Contains no named entities, which are important building blocks of fact or information. E.g., ``Check this out'', ``To say that a person is an example of something is absurd.''
    \vspace{-0.1em}
    \item Contains first-person pronouns (i.e., I, we, and us), which are strong signal for personal opinions or casual chitchat style of writing. E.g., ``I think...'', ``I believe...''
    \vspace{-0.1em}
    \item Contains question mark. E.g., ``Do you want to hear something interesting?'', ``Did you know?'', ``What are your thoughts?''
\vspace{-0.4em}
\end{itemize}

\section{Experiment Details}
\label{appendix:experiment_setup}
\paragraph{Usage Example of {\prefix}}
Here, we provide an example of how the training corpus looks like when {\prefix} is applied.

The following Wikipedia paragraph about Barack Obama:

\textit{Barack Hussein Obama II (born August 4, 1961) is an American politician who served as the 44th president of the United States from 2009 to 2017. He was the first African-American president of the United States. A member of the Democratic Party, he previously served as a U.S. senator from Illinois from 2005 to 2008 and as an Illinois state senator from 1997 to 2004.
}

is transferred into:

\textit{Barack Obama ==> Barack Hussein Obama II (born August 4, 1961) is an American politician who served as the 44th president of the United States from 2009 to 2017. Barack Obama ==> He was the first African-American president of the United States. Barack Obama ==> A member of the Democratic Party, he previously served as a U.S. senator from Illinois from 2005 to 2008 and as an Illinois state senator from 1997 to 2004.
}.

\paragraph{Hyper-parameters and training details}
The hyper-parameters for 1.3B factuality enhancement training were: learning rate 2e-6, batch size 64, maximum input sequence length 2048. For 530B model, the hyper-parameters were: learning rate 1e-5, batch size 512, maximum input sequence length 2048. 
The architecture details of pre-trained LMs are in Table \ref{tab:model_details}. During inference, we set maximum subword sequence length to be 150. \yeon{Same Wikipedia corpus with topic-prefix is commonly used for all our factuality enhancing training.}

\paragraph{Detail about pre-trained LMs}
\yeon{All LMs with different sizes are pre-trained on the same corpus, following the experimental details in \cite{shoeybi2019megatron}.}

\begin{table}[h]
\caption{ Architecture details of pre-trained LMs.}
\vspace{1mm}
\label{tab:model_details}
\small
\centering
{
\begin{tabular}{l|cccc}
\toprule
\multicolumn{1}{l|}{\multirow{1}{*}{\textbf{Models (\#/parameters)}}}  &  \#/layers & \#/hidden size & \#/ attention heads     \\
\midrule
126M & 12 & 768 & 12 \\
357M & 24 & 1024 & 16 \\
1.3B & 24 & 2048 & 32 \\
8.3B & 40 & 4096 & 64 \\
530B & 105 & 20480 & 128\\
\bottomrule
\end{tabular}
}
\end{table}
\vspace{-2mm}

\section{\prompts~Details}
\label{appendix:data_details}
\vspace{-2mm}
\begin{table}[h]
\centering
\small
\caption{Data statistics of \prompts}
\vspace{1mm}
\label{table:data_stat}
\begin{tabular}{rcc}
\toprule
\multicolumn{1}{l}{} & \textbf{Factual Prompts} & \textbf{Nonfactual Prompts} \\ \midrule
\# Prompts & 8000 & 8000 \\
Avg \# Tokens & 9.77 & 9.48 
\\ \bottomrule
\end{tabular}
\end{table}
Since high quality fact-related data collection requires a lot of human efforts, we instead utilize a well-established fact-related dataset, FEVER~\citep{thorne2018fever}, to construct our factual and nonfactual prompts.
FEVER is a fact-checking dataset consisting of claims that are \textsc{supported}, \textsc{refuted} or unverifiable (\textsc{NotEnoughInfo}) by Wikipedia documents. These claims are created by annotators who were asked to alter or paraphrase the sentences from Wikipedia. 
We leverage the \textsc{supported} and \textsc{refuted} claims from FEVER validation set~\footnote{The testset is not publicly released and can only be accessed through the FEVER workshop submission site. Therefore, it is common practice to leverage validation set instead.} as the factual and nonfactual prompts, respectively.
To further ensure the quality of the test set, we filter out claims that are not appropriate to serve as prompts -- e.g., extremely short claims that are not enough to provide any context to the LM.
The data statistics after filtering is reported in Table~\ref{table:data_stat}.

\section{\yeon{Limitations and Societal Impact}}
Although the factual-nucleus sampling requires the same amount of computation as regular top-$p$ sampling, the continued pre-training of large language models will have some negative carbon footprint. However, our task itself (trying to improve factuality) will bring more overall benefit to the community and society, by allowing the language models to generate less fake information and be safer for deployment. 
In terms of ethical consideration, to the best of our knowledge, Wikipedia has no private personal information or any inappropriate content (problematic discrimination towards particular demographic groups, NSFW contents, hate speech, etc). So, fine-tuning our model on it will not encourage unfairness, biases or toxic output.

\section{Extended Experimental Results}
\label{appendix:full_table_with_ppl}

\subsection{\yeon{Ablation Study of Sentence Completion Loss}}
A small scale experiments using 3000 Factual Prompts are conducted to explore the stand-alone impact of sentence completion loss. 
As shown in Table~\ref{table:sc_ablation}, the only having sentence completion loss is indifferent to having the standard factual-domain adaptive training (i.e., negligible difference in factuality). However, when used together with {\prefix}, it results in a significant boost for both factuality metrics.

\begin{table}[h!]
\caption{Ablation Study of Sentence Completion Loss}
\label{table:sc_ablation}
\centering
\begin{tabular}{lccc}
\toprule
\textbf{Model Choice}              & Decoding  & \halluNE$\downarrow$ & \entailedRatio$\uparrow$ \\ \midrule
\multirow{2}{*}{Default Wiki FT baseline}  & 0.9    & 55.92\%  & 2.58\%   \\ 
                          & 0.9 | 0.9 & 45.78\%  & 7.12\%           \\ \midrule
\multirow{2}{*}{\rootSC}  & 0.9       & 55.81\%  & 2.36\%           \\ 
                           & 0.9 | 0.9 & 44.80\%  & 6.39\%           \\ \midrule
\multirow{2}{*}{{\rootSC} + TopicPrefix}  & 0.9       & 35.15\%  & 6.67\%           \\ 
                           & 0.9 | 0.9 & 26.04\%  & 15.67\%          \\ \midrule
\multirow{2}{*}{\halfSC}  & 0.9       & 56.16\%  & 2.08\%           \\ 
                           & 0.9 | 0.9 & 45.13\%  & 6.64\%           \\ \midrule
\multirow{2}{*}{\halfSC~+ TopicPrefix}  & 0.9       & 34.18\%  & 7.63\%           \\ 
                           & 0.9 | 0.9 & 25.15\%  & 18.58\%          \\ \midrule
\multirow{2}{*}{\neSC}       & 0.9       & 56.42\%  & 1.88\%           \\ 
                           & 0.9 | 0.9 & 45.84\%  & 6.97\%           \\ \midrule
\multirow{2}{*}{{\neSC} + TopicPrefix}    & 0.9       & 35.87\%  & 4.59\%           \\ 
                           & 0.9 | 0.9 & 28.87\%  & 10.71\%          \\ \bottomrule
\end{tabular}
\end{table}

\subsection{Experimental Results with Perplexity}
In this subsection, we provide experimental results including the perplexity scores (PPL) of generated text evaluated on the 1.3B pretrained LM as a \emph{fluency} measure. The results consistently indicate that our proposed decoding and training methods do not harm the fluency of the generation. For instance, in Table~\ref{table:decoding_choices_w_ppl}, all our decoding choices result in PPL scores between $1.9\sim4.1$ that are smaller than Top-$p$ 0.9 PPL score $12.0$. 

To provide full details about the columns reported in Table~\ref{table:decoding_choices_w_ppl} and Table~\ref{table:factuality_enhanced_training_w_ppl}, \halluNE~refers to the named-entity error, \entailedRatio~refers to entailment ratio, Div. refers to distinct 4-grams and Rep. refers to repetition. $\uparrow$ means the higher the better, and $\downarrow$ means the lower the better.
 
\begin{table}[h!]
\caption{The factuality of \textbf{1.3B} LM with different decoding algorithms. $p$ is the nucleus probability, $\lambda$ is the decay factor, and $\omega$ lower bounds the decay. }
\label{table:decoding_choices_w_ppl}
\vspace{1mm}
\small
\begin{adjustbox}{width=1.01\textwidth}
\begin{tabular}{lccccc|ccccc}
\toprule
\multirow{2}{*}{\textbf{Decoding}} & \multicolumn{5}{c}{\textbf{Factual Prompt}} & \multicolumn{5}{c}{\textbf{Nonfactual Prompt}} \\ \cmidrule(lr){2-6} \cmidrule(lr){7-11}
 & \halluNE$\downarrow$ & \entailedRatio$\uparrow$ & \distFour$\uparrow$ & Rep.$\downarrow$ & PPL$\uparrow$ & \halluNE$\downarrow$ & \entailedRatio$\uparrow$ & \distFour$\uparrow$  & Rep.$\downarrow$ & PPL$\uparrow$ \\ \midrule
\emph{Greedy} & 39.9\% & 12.9\% & 0.05 & 33.1\% & 1.9 & 45.0\% & 8.8\% & 0.05 & 36.2\% & 2.0 \\ 
\textit{Top-p 0.9} & 52.4\% & 2.9\% & 0.88 & 0.2\% & 10.9 & 56.8\% & 2.0\% & 0.89 & 0.3\% & 12.0 \\
\midrule
\multicolumn{11}{l}{\ ~$p$ ~|~ $\lambda$ \hspace{.9cm}  Top-$p$ ~+~ \decay} \\
\midrule
0.9 | 0.9 & 41.1\% & 10.8\% & 0.43 & 30.7\% & 2.02 & 45.7\% & 6.8\% & 0.47 & 34.5\% & 2.13 \\
0.9 | 0.5 & 39.9\% & 13.0\% & 0.08 & 33.1\% & 1.89 & 44.9\% & 9.1\% & 0.09 & 35.9\% & 1.97 \\
\midrule
\multicolumn{11}{l}{\ ~$p$ ~|~ $\lambda$  \hspace{.9cm} Top-$p$ ~+~ \decay ~+~ $p$-reset } \\ \midrule
0.9 | 0.9 & 41.5\% & 10.3\% & 0.52 & 10.3\% & 3.6 & 45.4\% & 6.3\% & 0.57 & 9.1\% & 3.9 \\
0.9 | 0.5 & 39.3\% & 12.8\% & 0.34 & 17.8\% & 2.3 & 44.5\% & 8.4\% & 0.45 & 18.9\% & 2.5 \\
\midrule
\multicolumn{11}{l}{\ ~$p$ ~|~ $\lambda$ ~|~ $\omega$ \hspace{.3cm} Top-$p$ ~+~ \decay ~+~ \reset ~+~ \lowerBound~ (\emph{factual-nucleus sampling}) } \\ \midrule
0.9 | 0.9 | 0.3 & 42.1\% & 10.1\% & 0.55 & 7.1\% & 3.8 & 46.5\% & 5.6\% & 0.59 & 6.4\% & 4.1 \\
0.9 | 0.5 | 0.3 & 41.0\% & 12.2\% & 0.47 & 13.0\% & 2.8 & 46.0\% & 7.0\% & 0.51 & 12.7\% & 3.0 \\
0.9 | 0.9 | 0.2 & 41.7\% & 9.9\% & 0.52 & 8.6\% & 3.6 & 45.6\% & 6.2\% & 0.56 & 7.6\% & 4.0 \\
0.9 | 0.5 | 0.2 & 39.3\% & 12.8\% & 0.38 & 16.1\% & 2.5 & 45.2\% & 7.8\% & 0.42 & 16.9\% & 2.7 \\ 
\bottomrule
\end{tabular}
\end{adjustbox}
\end{table}

\begin{table}[h!]
\small
\centering
\caption{Results for factuality enhanced training. Decoding settings are formatted as: nucleus $p$ value, decay rate $\lambda$, lower-bound $\omega$}
\label{table:factuality_enhanced_training_w_ppl}
\vspace{1mm}
\begin{tabular}{lccccc|ccccc}
\toprule
\multirow{2}{*}{\textbf{Decoding}} & \multicolumn{5}{c}{\textbf{Factual Prompt}} & \multicolumn{5}{c}{\textbf{Nonfactual Prompt}} \\
\cmidrule(lr){2-6} \cmidrule(lr){7-11}
\textbf{($p$ | $\lambda$ | $\omega$)} & \multicolumn{1}{c}{\halluNE$\downarrow$} & \multicolumn{1}{c}{\entailedRatio$\uparrow$} & \multicolumn{1}{c}{\distFour} & \multicolumn{1}{c}{Rep.} & \multicolumn{1}{c}{PPL} & \multicolumn{1}{c}{\halluNE} & \entailedRatio & \distFour & Rep. & PPL \\ \midrule
\multicolumn{11}{l}{Vanilla Pretrained LM (1.3B) 
}\\ \midrule
0.9 & 52.4\% & 2.9\% & 0.88 & 0.2\% & 10.9 & 56.8\% & 2.0\% & 0.89 & 0.3\% & 12.0 \\
0.9 | 0.9 | 0.3 & 42.1\% & 10.1\% & 0.55 & 7.1\% & 3.8 & 46.5\% & 5.6\% & 0.59 & 6.4\% & 4.1 \\
\midrule
\multicolumn{11}{l}{Factual Domain-Adaptive Training with Wikipedia (1.3B)}\\ \midrule
0.9 & 52.5\% & 2.8\% & 0.85 & 0.2\% & 9.73 & 55.8\% & 2.2\% & 0.86 & 0.1\% & 10.69 \\
0.9 | 0.9 | 0.3 & 42.7\% & 7.1\% & 0.51 & 7.2\% & 3.60 & 48.2\% & 4.9\% & 0.56 & 6.0\% & 3.95 \\ \midrule
\multicolumn{11}{l}{{\prefix} (1.3B)} \\ \midrule
0.9 & 34.4\% & 4.2\% & 0.84 & 0.3\% & 8.03 & 36.2\% & 2.7\% & 0.85 & 0.2\% & 8.61 \\
0.9 | 0.9 | 0.3 & 27.6\% & 8.7\% & 0.43 & 8.0\% & 2.60 & 30.5\% & 6.1\% & 0.47 & 6.9\% & 2.75 \\ \midrule
\multicolumn{11}{l}{{\prefix} + \rootSC~(1.3B)}  \\ \midrule
0.9 & 32.5\% & 6.7\% & 0.83 & 1.2\% & 7.63 & 34.3\% & 4.6\% & 0.84 & 1.1\% & 8.15 \\
0.9 | 0.9 | 0.3 & 24.7\% & 15.8\% & 0.40 & 13.6\% & 2.32 & 27.6\% & 9.1\% & 0.44 & 13.7\% & 2.46 \\ \midrule
\multicolumn{11}{l}{{\prefix} + \randomSC~(1.3B)} \\ \midrule
0.9 & 32.0\% & 7.9\% & 0.81 & 1.2\% & 7.43 & 34.2\% & 5.5\% & 0.83 & 1.1\% & 7.96 \\
0.9 | 0.9 | 0.3 & 23.6\% & 17.6\% & 0.39 & 14.2\% & 2.31 & 26.9\% & 9.3\% & 0.42 & 13.2\% & 2.46 \\ \midrule
\multicolumn{11}{l}{{\prefix} + \halfSC~(1.3B)}  \\ \midrule
0.9 & 31.6\% & 7.6\% & 0.81 & 1.4\% & 7.33 & 33.5\% & 5.1\% & 0.83 & 1.5\% & 7.88 \\
0.9 | 0.9 | 0.3 & 23.6\% & 17.4\% & 0.38 & 14.4\% & 2.33 & 27.2\% & 10.2\% & 0.42 & 13.1\% & 2.47 \\ \midrule
\multicolumn{11}{l}{{\prefix} + \halfSC~(530B)} \\ \midrule
0.9 & 18.3\% & 19.3\% & 0.68 & 0.1\% & 8.30 & 21.7\% & 13.7\% & 0.68 & 0.1\% & 8.69 \\
0.9 | 0.9 | 0.3 & \textbf{14.5\%} & \textbf{25.5\%} & 0.33 & 0.2\% & 7.57 & \textbf{17.7\%} & \textbf{20.0\%} & 0.33 & 0.1\% & 7.71 \\
\bottomrule
\end{tabular}
\end{table}

\clearpage
\section{Generation Examples (Continued)}
\label{appendix:more_generations_2}

\begin{table}[h!]
\small
\centering
\begin{tabular}{p{0.95\linewidth}}
\toprule
\textbf{[Factual Prompt]} The ruins of the ancient Roman town of Pompeii lie near Naples \\ \midrule
\textbf{[Greedy]} The city was destroyed by a volcanic eruption in 79 AD.\textbackslash{}n\textbackslash{}nThe ruins of the ancient Roman town of Pompeii lie near Naples. The city was destroyed by a volcanic eruption in 79 AD.\textbackslash{}n\textbackslash{}nPhoto: AP\textbackslash{}n\textbackslash{}nImage 2 of 3\textbackslash{}n\textbackslash{}nA man walks past a mural depicting the ancient Roman town of Pompeii in Naples, Italy.\textbackslash{}n\textbackslash{}nA man walks past a mural depicting the ancient Roman town of Pompeii in Naples, Italy.\textbackslash{}n\textbackslash{}nPhoto: AP\textbackslash{}n\textbackslash{}nImage 3 of 3\textbackslash{}n\textbackslash{}nPompeii's ruins to be restored\textbackslash{}n\textbackslash{}n1 / 3\textbackslash{}n\textbackslash{}nBack to Gallery\textbackslash{}n\textbackslash{}nROME 
\\
\vspace{0mm}
\textbf{[p=0.9]} Millions of tourists from around the world come to visit them each year. They enter through a huge gate. This gate was built when Pompeii became a part of Italy again in the nineteenth century. Before that time, no one was allowed to enter the area.\textbackslash{}n\textbackslash{}nBut that is not the only gate through which visitors enter Pompeii.\textbackslash{}n\textbackslash{}nThe back of the entrance gate is very high. For a long time, no one has known why it is so tall. It seemed as if the builder simply did not finish it.\textbackslash{}n\textbackslash{}nOne day a
\\
\vspace{0mm}
\textbf{[Ours]} Naples has a wide range of different architectural styles , from the medieval castles in the historic centre to the Art Nouveau buildings in the \" Vomero \" area , to the buildings of the 20th century , such as the San Carlo opera house . \textbackslash{}nNaples is also widely known for its wealth of historical museums . \textbackslash{}nThe Naples National Archaeological Museum is one of the city 's main museums , with one of the most extensive collections of artefacts of the Roman Empire in the world . \textbackslash{}nIt also houses many of the antiques unearthed at Pompeii and Herculaneum , as well as some artefacts from the Greek and Renaissance periods . \textbackslash{}nPreviously a Bourbon palace , now a museum and art gallery , the Museo di Capodimonte is another" \\
\midrule \midrule 

\textbf{[Factual Prompt]} David Beckham is a player for the Major League Soccer club LA Galaxy. \\ \midrule
\textbf{[Greedy]} He is also a UNICEF ambassador.\textbackslash{}n\textbackslash{}nThe opinions expressed in this commentary are solely those of David Beckham.<|endoftext|>
\\
\vspace{0mm}
\textbf{[p=0.9]} For England, he was a very successful player, playing for Man U (again and again), and captaining the team for several years, until he fell out of favor with his coach.\textbackslash{}n\textbackslash{}nIn 2003, after a free kick from England's David Beckham, Argentinean player and infamous cokehead Diego Simeone headbutted Beckham. According to Beckham, Simeone leered at him and said "I'm going to fuck you." Simeone claimed the players collided accidentally, but the ref gave him a red card for the incident, ...
\\
\vspace{0mm}
\textbf{[Ours]} He is the first English player to win league titles in four countries : England , Spain , the United States and France . He retired in May 2013 after a 20-year career , during which he won 19 major trophies . Beckham 's professional club career began with Manchester United , where he made his first - team debut in 1992 at age 17 . With United , he won the Premier League title six times , the FA Cup twice , and the UEFA Champions League in 1999 . 
He then played four seasons with Real Madrid , winning the La Liga championship in his final season with the club . 
\\
\midrule \midrule 

\textbf{[Factual Prompt]} Andy Warhol has creations that are highly valuable. \\ \midrule
\textbf{[Greedy]} His works are sold for millions of dollars. He is one of the most famous artists in the world.\textbackslash{}n\textbackslash{}nAndy Warhol was born in Pittsburgh, Pennsylvania. He was the youngest of three children. His parents were immigrants from Czechoslovakia. His father was a construction worker. His mother was a homemaker.\textbackslash{}n\textbackslash{}nAndy Warhol was a shy child. He was often sick. He had a nervous stomach. He was often in the hospital. He was afraid of the dark. He was afraid of the dentist. He was afraid of doctors. He was afraid of hospitals.\textbackslash{}n\textbackslash{}nAndy Warhol went to college. He studied art. He was a good student. He was a good artist. He was a good painter. He was a good drawer. He was a good illustrator. He was a good graphic artist...
\\
\vspace{0mm}
\textbf{[p=0.9]} One of these creations is a book. During his days as an art student, Warhol used to paint advertisements that he printed on pieces of cardboard. He pasted these advertisements on a cork and this is what he called his book. For Warhol, making a book meant simply arranging his images on a page. He didn't make the necessary effort to bind the pages together to make it qualify as a book. This was the same way Warhol created many of his creations and works. He could have made every page of his work a masterpiece, but he didn't do...
\\
\vspace{0mm}
\textbf{[Ours]} Warhol was an inspiration towards Cattelan 's magazine and photography compilations , such as " Permanent Food , Charley " , and " Toilet Paper " . \textbackslash{}nIn the period just before his death , Warhol was working on "Cars", a series of paintings for Mercedes - Benz . \textbackslash{}nA self - portrait by Andy Warhol ( 1963-64 ) , which sold in New York at the May Post - War and Contemporary evening sale in Christie 's , fetched \$ 38.4 million . \textbackslash{}nOn May 9 , 2012 , his classic painting " Double Elvis ( Ferus Type ) " sold at auction at Sotheby 's in New York for US\$ 33 million . \textbackslash{}nWith commission , the sale price totaled US\$ 37,042...
\\
\bottomrule
\end{tabular}
\end{table}

\clearpage
\begin{table}[h!]
\small
\centering
\begin{tabular}{p{0.95\linewidth}}
\toprule
\textbf{[Factual Prompt]} San Francisco is the location of a bridge. \\ \midrule
\textbf{[Greedy]} The bridge is located in San Francisco.\textbackslash{}n\textbackslash{}nThe bridge is located in San Francisco.\textbackslash{}n\textbackslash{}nThe bridge is located in San Francisco.\textbackslash{}n\textbackslash{}nThe bridge is located in San Francisco.\textbackslash{}n\textbackslash{}nThe bridge is located in San Francisco.\textbackslash{}n\textbackslash{}nThe bridge is located in San Francisco.\textbackslash{}n\textbackslash{}nThe bridge is located in San Francisco.\textbackslash{}n\textbackslash{}nThe bridge is located in San Francisco.\textbackslash{}n\textbackslash{}nThe bridge is located in San Francisco.\textbackslash{}n\textbackslash{}nThe bridge is located...
\\
\vspace{0mm}
\textbf{[p=0.9]} San Francisco is an island surrounded by bridges.\textbackslash{}n\textbackslash{}nOnce across the bridges, there is much to explore. Chinatown, for example, has interesting shops, food, and people-watching. Walk in Chinatown and you will hear many languages, see many signs in different scripts, and smell interesting things from kitchens. From Chinatown you can see the world.\textbackslash{}n\textbackslash{}nBut Chinatown is not what this is about. What this is about is part of San Francisco which is on the east side of the city, about three miles from Chinatown...
\\
\vspace{0mm}
\textbf{[Ours]} There are seven major islands in the San Francisco Bay : Alameda , Marin , Angel , Treasure , Yerba Buena , Contra Costa and Alcatraz . Alcatraz , in the middle of the bay , is a federal penitentiary which also housed a lighthouse for many years . It is now a public museum and one of San Francisco 's major tourist attractions , attracting some 1.5 million visitors annually . Yerba Buena Island , between San Francisco and Oakland , contains the San Francisco Port of Embarkation , the Oakland Naval Supply Depot , and a Coast Guard station ...
\\
\midrule \midrule

\textbf{[Factual Prompt]} Michael Jordan was in "The Dream Team". \\ \midrule
\textbf{[Greedy]} The Dream Team was the nickname of the USA men's basketball team that won the gold medal at the 1992 Olympics.\textbackslash{}n\textbackslash{}nThe Dream Team was the nickname of the USA men's basketball team that won the gold medal at the 1992 Olympics.\textbackslash{}n\textbackslash{}nThe Dream Team was the nickname of the USA men's basketball team that won the gold medal at the 1992 Olympics.\textbackslash{}n\textbackslash{}n
\\
\vspace{0mm}
\textbf{[p=0.9]} Mickey Mouse is Disney.\textbackslash{}n\textbackslash{}nSophia Loren was an actress from Italy.\textbackslash{}n\textbackslash{}nLast but not least is Harry Potter.\textbackslash{}n\textbackslash{}nAll these brands immediately create a strong visual image in the mind of the target audience, and this is the same for the faith-based ministry.\textbackslash{}n\textbackslash{}nTo be seen as an iconic faith-based ministry the church must have a look and feel that is instantly recognizable.\textbackslash{}n\textbackslash{}nDo you have a recognizable logo?\textbackslash{}n\textbackslash{}nDoes your website and social media pages look like the brand of your church?\textbackslash{}n\textbackslash{}nDo you have
\\
\vspace{0mm}
\textbf{[Ours]} In the summer of 1992 , Jordan led the Dream Team to the Olympic gold medal in Barcelona , scoring a game - high 26.5 points per game , and making a game - winning jump shot against Croatia . Jordan was named the tournament 's MVP , although the Olympic basketball competition was later considered to be an exhibition for NBA players , and thus NBA players were not allowed to compete in future Olympic basketball tournaments . In March 1993 , Jordan announced his retirement , citing a loss of desire to play the game . Jordan later stated that the death of his father three months earlier also shaped his decision . Jordan 's father was murdered on July 23 , 1993 , at a highway rest area in Lumberton , North Carolina , by two teenagers , Daniel Green and Larry Martin Demery , who carjacked his luxury Lexus bearing the license plate \" UNC 0023 \" . His body was dumped in a South Carolina swamp and was not discovered until August 3 . 
\\
\midrule \midrule 

\textbf{[Factual Prompt]} Cara Delevingne had a role in a film. \\ \midrule
\textbf{[Greedy]} \textcolor{black}{Cara Delevingne had a role in a film. Cara Delevingne had a role in a film. Cara Delevingne had a role in a film. Cara Delevingne had a role in a film. Cara Delevingne had a role in a film. Cara Delevingne had a role in a film. Cara Delevingne had a role in a film. Cara Delevingne had a role in a film.} \\
\vspace{0mm}
\textbf{[p=0.9]}
\textcolor{black}{She's actually one of the many faces in a sketch for a film by celebrated photographer David LaChapelle called "Lost and Found."} And I'm not going to lie to you, it's kind of terrifying, but in a good way? I think.<|endoftext|>" \\
\vspace{0mm}
\textbf{[Ours]} \textcolor{black}{Delevingne's first film appearance was in the 2012 film adaptation of "Anna Karenina", as Princess Sorokina, the marriage interest of Count Vronsky, alongside Keira Knightley. In August 2013, Delevingne voiced a DJ of a pop radio station in the video game "Grand Theft Auto V," which once released became the fastest - selling entertainment product in history.} \\

\bottomrule
\end{tabular}
\end{table}

\clearpage
\begin{table}[h!]
\small
\centering
\begin{tabular}{p{0.95\linewidth}}
\toprule
\textbf{[Factual Prompt]} Doctor Who is a show on television.
 \\ \midrule
\textbf{[Greedy]} \textcolor{black}{It is a science fiction show. It is a show about a man who travels through time and space in a blue box.} \textcolor{black}{It is a show about a man who is over 900 years old.} \textcolor{black}{It is a show about a man who is a Time Lord. It is a show about a man who is a Time Lord who has a time machine called the TARDIS. It is a show about a man who is a Time Lord who has a time machine called the TARDIS that is bigger on the inside than it is on the outside. It is a show about a man who is a Time Lord who has a time machine called the TARDIS that is bigger on the inside than it is on the outside and that is shaped like a police box.}
\\
\vspace{0mm}
\textbf{[p=0.9]} The men on the show are the Doctor and the Master. The women are the companions, and they are held captive until they are rescued, or dead. When they are rescued, it's rarely by the Doctor. Even when the Doctor is in the scene, the female characters are passively waiting to be saved. Here are a few examples.
\textbackslash{}n\textbackslash{}nSeason 5. Episode 13. "The Lodger"\textbackslash{}n\textbackslash{}nhttp:\/\/www.youtube.com\/watch?v=bHNce0\_Qlrk\textbackslash{}n\textbackslash{}nHere's a fun little game.
\\
\vspace{0mm}
\textbf{[Ours]} Doctor Who is a British science fiction television programme produced by the BBC since 1963 . The programme depicts the adventures of a Time Lord called " the Doctor " , an extraterrestrial being , to all appearances human , from the planet Gallifrey . The Doctor explores the universe in a time - travelling space ship called the TARDIS . Its exterior appears as a blue British police box , which was a common sight in Britain in 1963 when the series first aired . Accompanied by a number of companions , the Doctor combats a variety of foes while working to save civilisations and help people in need . The show is a significant part of British popular culture , and elsewhere it has gained a cult following...
\\ \midrule \midrule

\textbf{[Factual Prompt]} Tiber Oil Field is operated by BP. \\ \midrule
\textbf{[Greedy]} See also List of oil fields References Category:Oil fields in Kern County, California\textbackslash{}nCategory:Geography of the San Joaquin Valley\textbackslash{}nCategory:BP oil and gas fields<|endoftext|>
\\
\vspace{0mm}
\textbf{[p=0.9]} The rig is located approximately off the Louisiana coast. It is a riserless subsea design, with a maximum structural height. This is the first use of this design. Operations Production on the field was expected to begin in 2011, but first oil was delayed until 2014.  Oil is produced through a wellhead located on the seafloor, and it is then piped up to a floating production storage and offloading (FPSO) vessel. Oil from this vessel is then transported to shore. The project is being developed in...
\\
\vspace{0mm}
\textbf{[Ours]} Tiber Oil Field is located in the Keathley Canyon block 102 of the United States sector of the Gulf of Mexico . Tiber is only one of four deepwater fields in the Gulf of Mexico that is operated by BP . The oil field was discovered in September 2009 and it is located in Keathley Canyon block 102 about southeast of Houston and south west of New Orleans . Tiber is BP 's first ultra - deepwater platform in the Gulf of...
\\ \midrule \midrule
\textbf{[Factual Prompt]} The Blue Lagoon was released in or before 1930. \\ \midrule
\textbf{[Greedy]} The Blue Lagoon was released in or before 1930. The Blue Lagoon was released in 1931.\textbackslash{}n\textbackslash{}nThe Blue Lagoon was released in 1932.\textbackslash{}n\textbackslash{}nThe Blue Lagoon was released in 1933.\textbackslash{}n\textbackslash{}nThe Blue Lagoon...
\\
\vspace{0mm}
\textbf{[p=0.9]} The film was re-released on 5 July 1949, with a higher age restriction, and three minutes of footage from the previous cut removed. The 1949 film was retitled The Island of Purity. The film was released on DVD on 6 Jan 2004...
\\
\vspace{0mm}
\textbf{[Ours]} The novel has inspired several film adaptations, most notably "The Blue Lagoon" starring Brooke Shields as Emmeline and Christopher Atkins as Richard ("Dicky" in the book), in which they are two fictional English children. The story...
\\
\midrule \midrule 
\textbf{[Factual Prompt]} Minos had a wife and he was Greek. \\ \midrule
\textbf{[Greedy]} The Greek was a man of many words.\textbackslash{}n\textbackslash{}nThe Greek was a man of many words.\textbackslash{}n\textbackslash{}nThe Greek was a man of many words.\textbackslash{}n\textbackslash{}nThe Greek was a man of many words.\textbackslash{}n\textbackslash{}nThe Greek was a man of many words.\textbackslash{}n\textbackslash{}nThe Greek was a man of many words.\textbackslash{}n\textbackslash{}nThe Greek was a man of many words.\textbackslash{}n\textbackslash{}nThe Greek was a man of many words.\textbackslash{}n\textbackslash{}nThe... 
\\
\vspace{0mm}
\textbf{[p=0.9]} For this reason, this creature did not consider himself to be a Satan.\textbackslash{}n\textbackslash{}nEros was all about love.\textbackslash{}n\textbackslash{}nHe loved his wife and didn't want to hurt her.\textbackslash{}n\textbackslash{}nFor this reason, he also didn't believe he was an angel.\textbackslash{}n\textbackslash{}nLove did not originate from Hell.  It's original abode was Heaven. It was only much later, when his job was downsized, he became to earn his living by showing his love through sex. This is why today, love is known to be
\\
\vspace{0mm}
\textbf{[Ours]} Minos was the son of Zeus and Europa. He lived at Knossos, where he received the laws from Zeus himself. He was the founder and first king of the Minoan dynasty. He was also the judge of the underworld. In Crete , he was the founder - king of the Minoan civilization and the palace of Knossos was the centre of the Minoan civilization . On the Greek mainland , he was the founder...
\\
\bottomrule
\end{tabular}
\end{table}

\end{document}